\documentclass[lettersize,journal]{IEEEtran}
\usepackage{amsmath,amsfonts}
\usepackage{algorithm}
\usepackage{algpseudocode}
\usepackage{array}
\usepackage[caption=false,font=normalsize,labelfont=sf,textfont=sf]{subfig}
\usepackage{textcomp}
\usepackage{stfloats}
\usepackage{url}
\usepackage{verbatim}
\usepackage{graphicx}
\usepackage{cite}
\usepackage{multirow}
\usepackage{booktabs}
\usepackage{amsmath}
\usepackage[table]{xcolor}
\usepackage{amssymb}
\usepackage{bm}
\usepackage{bbm}
\usepackage{utfsym}

\begin{document}

\title{
MambaTAD: When State-Space Models Meet \\ Long-Range Temporal Action Detection
}

\author{
Hui Lu, Yi Yu, Shijian Lu, Deepu Rajan,~\IEEEmembership{Member,~IEEE,}
Boon Poh Ng, Alex C. Kot,~\IEEEmembership{Life Fellow,~IEEE,} \\
Xudong Jiang,~\IEEEmembership{Fellow,~IEEE}
\IEEEcompsocitemizethanks{
        \IEEEcompsocthanksitem Hui Lu and Yi Yu are with the Rapid-Rich Object Search Lab, Interdisciplinary Graduate Programme, Nanyang Technological University, Singapore, (e-mail: \{hui007, yuyi0010\}@e.ntu.edu.sg).
  \IEEEcompsocthanksitem Shijian Lu and Deepu Rajan are with the College of Computing and Data Science, Nanyang Technological University, Singapore, (e-mail: \{shijian.Lu, asdrajan\}@ntu.edu.sg).
  \IEEEcompsocthanksitem Boon Poh Ng, Alex C. Kot, and Xudong Jiang are with the School of Electrical and Electronic Engineering, Nanyang Technological University, Singapore, (e-mail: \{ebpng, eackot, exdjiang\}@ntu.edu.sg).

 }
\thanks{
This work was carried out at the Rapid-Rich Object Search (ROSE) Lab, School of Electrical and Electronic Engineering, Nanyang Technological University (NTU), Singapore.
 (Corresponding author: Yi Yu.)} 
}

\markboth{Journal of \LaTeX\ Class Files,~Vol.~14, No.~8, August~2021}%
{Shell \MakeLowercase{\textit{et al.}}: A Sample Article Using IEEEtran.cls for IEEE Journals}


\maketitle

\begin{abstract}
Temporal Action Detection (TAD) aims to identify and localize actions by determining their starting and ending frames within untrimmed videos. Recent Structured State-Space Models such as Mamba have demonstrated potential in TAD due to their long-range modeling capability and linear computational complexity. On the other hand, structured state-space models often face two key challenges in TAD, namely, decay of temporal context due to recursive processing and self-element conflict during global visual context modeling, which become more severe while handling long-span action instances. 
Additionally, traditional methods for TAD struggle with detecting long-span action instances due to a lack of global awareness and inefficient detection heads.
This paper presents MambaTAD, a new state-space TAD model that introduces long-range modeling and global feature detection capabilities for accurate temporal action detection. 
MambaTAD comprises two novel designs that complement each other with superior TAD performance. First, it introduces a Diagonal-Masked Bidirectional State-Space (DMBSS) module which effectively facilitates global feature fusion and temporal action detection. Second, it introduces a global feature fusion head that refines the detection progressively with multi-granularity features and global awareness. 
In addition, MambaTAD tackles TAD in an end-to-end one-stage manner using a new state-space temporal adapter(SSTA) which reduces network parameters and computation cost with linear complexity. 
Extensive experiments show that MambaTAD achieves superior TAD performance consistently across multiple public benchmarks.
\end{abstract}


\section{Introduction}
\IEEEPARstart{T}emporal action detection (TAD) aims to detect specific action categories and extract corresponding temporal spans in untrimmed videos. It is a long-standing and challenging problem in video understanding with extensive real-world applications such as sports analysis, surveillance and security. The development of deep neural networks such as CNNs \cite{shi2023tridet, yang2024dyfadet} and Transformers \cite{tan2021relaxed, yang2024adapting} has led to continuous advancements in TAD performance over the past few years. However, CNNs have limited capabilities in capturing long-range dependencies, while Transformers face challenges with computational complexity and feature discrimination~\cite{shi2023tridet}.

\begin{figure}[t]
\centering
\includegraphics[width=0.75\columnwidth]{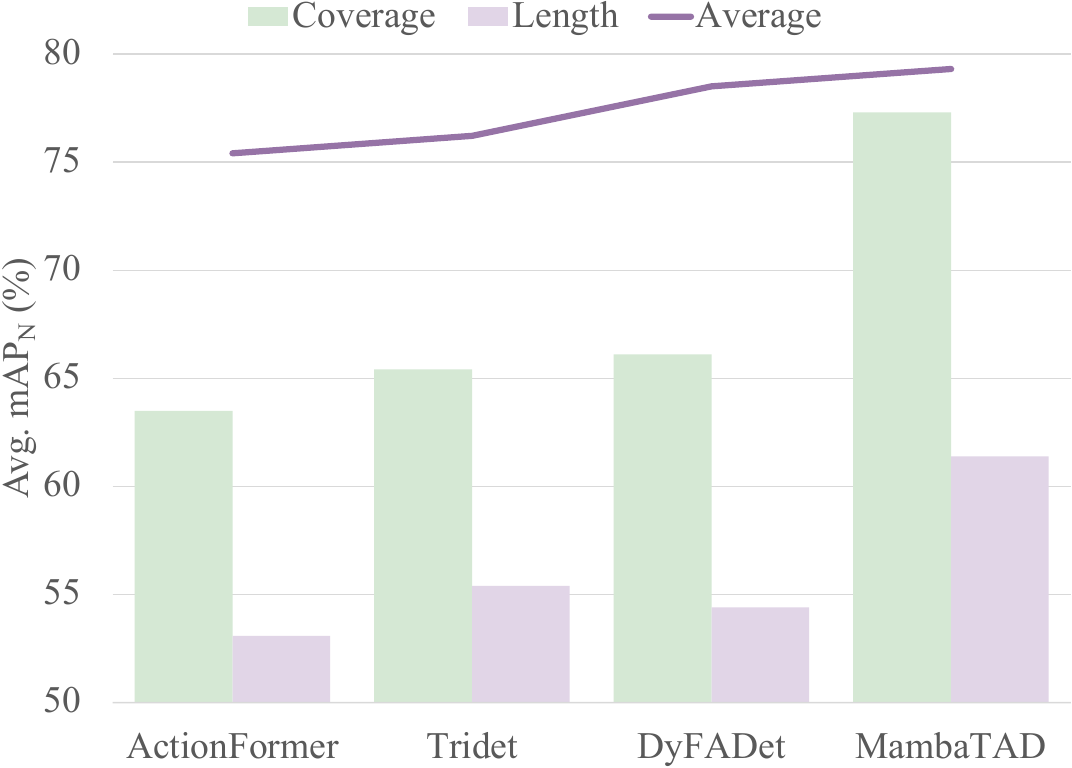} 
\caption{Comparison of TAD methods: Prior studies suffer from decay of temporal information and self-element conflict, which often struggle while facing long-span action instances. The proposed MambaTAD can handle long-span action instances effectively with its long-range modeling and global feature fusion capabilities. The \textit{Coverage} and \textit{Length} are two metrics for identifying long action instances according to their proportion with respect to the whole videos ([0.08,1]) and the absolute action length ([18,$\infty$] seconds), respectively. The \textit{Average} means the normalized average mAP over all action instances of various lengths in the dataset.
}
\label{fig1}
\end{figure}

Recently, Structured State-Space Sequence models (S4)~\cite{gu2021efficiently} such as Mamba \cite{gu2023mamba} have demonstrated great efficiency and effectiveness in deep network construction~\cite{DBLP:conf/icml/ZhuL0W0W24,ma2024u}. These models, enhanced by specially designed structured re-parameterization \cite{gu2020hippo} and selective scan mechanisms, facilitate the natural activation of extended temporal moments, thereby improving classification and boundary regression performance. However, the standard Mamba, which is designed for long sequence data in natural language using one forward branch, is not a natural fit for the TAD task. Specifically, Mamba processes flattened one-dimensional sequences in a recursive manner. It often loses temporal information of earlier moments and suffers from the problem of decay of temporal information due to the involved lower triangular matrices \cite{dao2024transformers}. 
In addition, since the trainable weights are the incorporation of a lower triangular matrix and an upper triangular matrix in bidirectional Mamba \cite{li2024videomamba, chen2024video}, they often face the problem of self-element conflict (\textit{i.e.}, diagonal conflict) while modeling global visual context. Such self-element conflict impedes the model from learning critical temporal boundary representations effectively.

These observations motivate us to present MambaTAD, a novel state-space model for effectively handling the challenging TAD problem. MambaTAD consists of a detector including two novel designs and an adapter for the backbone. The first is a pyramid model featuring a Diagonal-Masked Bidirectional State-Space (DMBSS) module. The module introduces a dual-branch input inversion mechanism that mitigates the decay of temporal information better and preserves global temporal information by retrieving earlier information via input flipping. This directly helps capture the progression and structure of long-duration actions. On top of that, the model mitigates diagonal conflict by masking the diagonal elements in backward matrices. The second is a global feature fusion head that integrates features of different granularities to enhance global awareness, enabling the model to capture both fine-grained details and broad patterns like slow motion in actions. At the instance level, it means that the model can recognize small, fast movements as well as the overall structure and flow of the action across time. As in Figure \ref{fig1}, MambaTAD outperforms DyFADet \cite{yang2024dyfadet} by 11.2\% and 7.0\% in mAP for long-duration actions under the \textit{Coverage} and \textit{Length} categories, respectively. 
 
 What is more, end-to-end training in TAD involves simultaneously optimizing both the video encoder and the action detector as a unified model. To further extend the capability of DMBSS beyond the detector, we propose the State-Space Temporal Adapter (SSTA), a novel parameter-efficient module designed for end-to-end TAD. Unlike conventional fine-tuning approaches that require extensive parameter updates\cite{ zhao2023re2tal,yang2024adapting}, SSTA leverages DMBSS within a state-space modeling framework, enabling efficient adaptation while preserving global and local temporal dependencies.
 Extensive experiments over five TAD datasets show that MambaTAD outperforms SOTA methods consistently with fewer parameters and lower FLOPs.

 The main contributions of this work can be summarized as:
\begin{enumerate}
    \item We propose a unified end-to-end one-stage framework that introduces state-space models for effective end-to-end temporal action detection. To the best of our knowledge, this is the first attempt at end-to-end TAD.
    \item We design two novel modules for effective TAD, including a Diagonal-Masked Bidirectional State-Space (DMBSS) module which solves problems of decay of temporal context and diagonal conflict when using vanilla Mamba for joint and effective action detection and action boundary localization with fewer parameters and lower FLOPs and a global feature fusion head for better global awareness by integrating features of different pyramid layers. 
    \item We propose an efficient state-space temporal adapter to adapt the humongous backbone as well as aggregate the global temporal context for end-to-end TAD. 
    \item Extensive experiments show that MambaTAD outperforms SOTA methods across five TAD datasets. 
\end{enumerate}

\section{Related Work}
\subsection{Temporal Action Detection}
TAD methods are broadly classified as two-stage and one-stage approaches. Two-stage methods \cite{qing2021temporal, lin2019bmn, liu2019multi} involve proposal generation and classification but face challenges due to high complexity, limiting end-to-end training. Recent trends favor one-stage frameworks, which support end-to-end optimization.  
One-stage methods are further divided into anchor-based and anchor-free approaches. Anchor-based methods \cite{chen2019relation, chen2020afnet, lin2017single, long2019gaussian} assign multi-scale intervals to temporal locations, while anchor-free methods \cite{lin2021learning, xia2023exploring, zhao2022temporal} generate flexible proposals with precise boundaries. MambaTAD falls into the one-stage anchor-free category.  
Different methods employ CNNs \cite{yang2023basictad, yang2020revisiting, liu2018deep, yan2020higcin}, Graph models \cite{zhao2021video, zeng2019graph}, Transformers \cite{liu2022end, 10440542, yan2023progressive, su2023sequence}, or Diffusion models \cite{nag2023difftad} for temporal encoding, whereas our approach is based on state-space models.

TAD methods are also classified into feature-based and end-to-end approaches. Feature-based methods use pre-extracted RGB and optionally optical flow features, while end-to-end methods process raw video frames, jointly optimizing the video encoder and action detector\cite{liu2020progressive}. E2E-TAD \cite{liu2022empirical} analyzed design choices affecting TAD performance and proposed a mid-resolution baseline due to GPU constraints.  
Advancements like TALLFormer \cite{cheng2022tallformer}, which introduced partial data backpropagation, and Re2TAL \cite{zhao2023re2tal}, which leveraged reversible architectures for memory efficiency, have much improved TAD. AdaTAD \cite{liu2024end} further scaled TAD to 1 billion parameters via PEFT. To our knowledge, we are the first to explore end-to-end TAD in state-space models.
\subsection{State-Space Models for Long-range Modeling}
State-Space Models (SSMs) have gained attention for their efficient linear scaling in long-range dependency modeling. The Structured State-Space Sequence model \cite{gu2021efficiently} pioneered this area, followed by the S5 layer \cite{smith2022simplified}, which introduced MIMO SSM and parallel scanning. Advances like H3 \cite{fu2022hungry} narrowed the gap with Transformers, while the Gated State Space layer \cite{mehta2022long} enhanced S4 with gating. Recently, Mamba \cite{gu2023mamba} introduced a selective mechanism and optimized hardware design, surpassing Transformers in large-scale tasks while preserving linear complexity. Mamba2~\cite{dao2024transformers} further improves via viewing Transformers as structured state space models, enabling efficient algorithms via state space duality, while Hydra~\cite{hwang2024hydra} uses bidirectional SSMs with matrix mixers to better capture past and future context. Our approach is developed to tackle the challenges of applying Mamba in TAD and, in principle, can be extended to a wide range of foundational state-space models.
\subsection{Visual State-Space Models}
Several studies have applied SSMs to various visual foundation models or developed hybrid architectures by combining SSMs with convolution or attention mechanisms. For instance, \cite{DBLP:conf/icml/ZhuL0W0W24, li2024videomamba, lu2024videomambapro} 
 integrate bidirectional SSMs for data-dependent global visual context modeling and employ position embedding to enhance location-aware visual understanding. Moreover, many pioneering works have applied visual state-space models to visual tasks, like \cite{islam2022long} 
 utilized 1D S4 to manage long-range temporal dependencies 
 for video classification, \cite{ma2024u} 
 proposes a hybrid CNN-SSM architecture to handle the long-range dependencies 
for biomedical image segmentation, \cite{mondal2024hummuss} for human motion understanding, and
 processes human motion using SSMs, 
 \cite{guo2024mambair} 
 explores the potential of Mamba to 
 for image restoration. Recently, CausalTAD~\cite{liu2024harnessing} further leverages temporal causality in SSMs to improve temporal action detection by integrating with causal attention blocks, while our DMBSS remains purely SSM-based and lightweight. \cite{chen2024video} proposed a Decomposed Bidirectionally Mamba (DBM) block for many video understanding tasks, including temporal action detection. 
 Our approach further addresses the limitations of Mamba in TAD while introducing the DMBSS and a global awareness detection head, enhancing flexibility and efficiency in modeling both action and boundary representations.
 
\section{Methodology}
\subsection{Preliminaries}
Recent progress in the development of SSMs, \textit{e.g.} S4 and Mamba, is largely inspired by Continuous Linear Time-Invariant (LTI) system theory. These models transform a one-dimensional input sequence $x(t) \!\in\! \mathbb{R}$ into an output $y(t) \in \mathbb{R}$ through a high-dimensional latent state $h(t) \in \mathbb{R}^n$, where $n$ is the size of the state. Their dynamics are governed by linear Ordinary Differential Equations (ODEs), capturing the temporal evolution of the latent state.
\begin{equation}\small\small
    h^\prime(t)  = \mathbf{A}\cdot h(t) + \mathbf{B}\cdot x(t),~
    y(t)  = \mathbf{C}\cdot h(t),
  \label{eqn-1}
\end{equation}
where $\mathbf{A} \in \mathbb{R}^{n \times n}$, $\mathbf{B} \in \mathbb{R}^{n \times 1}$, $\mathbf{C}\in\mathbb{R}^{1\times  n}$ are trainable parameters, named state space, of neural networks. To address the discrete input sequence $\mathbf{x}=(x_0,x_1,\ldots, x_{s-1})\in\mathbb{R}^s$,  S4 employs a step size, denoted as $\Delta$, to quantize the parameters in Eq. \ref{eqn-1}. This step size effectively determines the granularity or resolution at which the continuous input $x(t)$ is represented. Notably, the continuous variables $\mathbf{A, B}$ are transformed into their discrete counterparts through the application of a Zero-Order Hold (ZOH) technique given by
\begin{equation}\small\small
    \bar{\mathbf{A}}=exp(\Delta \mathbf{A}),~
    \bar{\mathbf{B}}=(\Delta \mathbf{A})^{-1}\cdot[exp(\Delta \mathbf{A})-\mathbf{I}]\cdot\Delta\mathbf{B}.
  \label{eqn-2}
\end{equation}
After the discretization process with a step size of $\Delta$, the discrete representation of Eq.~\ref{eqn-1} can be reformulated into the recurrent neural network (RNN) format as follows: 
\begin{equation}  \small\small
    h_t  = \bar{\mathbf{A}}\cdot h_{t-1} + \bar{\mathbf{B}}\cdot x_t,~
    y_t  = \mathbf{C}\cdot h_t.
  \label{eqn-3}
\end{equation}
To improve computational efficiency and ensure scalability, the iterative Eq. \ref{eqn-3} can be effectively consolidated using a global convolution operation. This approach allows for the parallel processing of data, leading to enhanced performance and better management of large-scale computations: 
\begin{equation}\small\small
    \bar{\mathbf{K}}=(\mathbf{C} \bar{\mathbf{B}}, \mathbf{C} \bar{\mathbf{A}}\bar{\mathbf{B}}, \ldots, \mathbf{C} \bar{\mathbf{A}}^{s-1}\bar{\mathbf{B}}),~
    \mathbf{y}=\mathbf{x} \ast \bar{\mathbf{K}},
  \label{eqn-4}
\end{equation}
where the variable $\bar{\mathbf{K}}\in \mathbb{R}^s$ corresponds to a structured convolution kernel derived from SSMs, $s$ is the length of the input sequence, and $\ast$ is the convolution operation. 
The cutting-edge state-space model, Mamba \cite{gu2023mamba} proposed a selective scan mechanism, which refined elements $\mathbf{B, C}$ and $\Delta$ to become input-adaptive, thereby enabling a more dynamic feature representation:
\begin{equation}\small\small
    \mathbf{B, C}=Linear(\mathbf{x}),~
    \Delta=softplus(\mathbf\Delta+Linear(\mathbf{x})),
  \label{eqn-5}
\end{equation}
where $\mathbf{B,C}\in \mathbb{R}^{s\times n}$ and $\mathbf\Delta ^{s\times c}$ are dependent on the input sequence $\mathbf{x}$. In Mamba, the scalar step size $\Delta$ becomes input-dependent and is broadcast to a tensor $\mathbf{\Delta}$. Here, $s$ is the sequence length, $c$ is the number of channels, and $n$ is the state size. 

\noindent\textbf{Problem definition.}
MambaTAD is a unified end-to-end one-stage anchor-free and attention-free architecture that processes raw untrimmed videos or temporal video features as input and produces representations for each action category and their temporal localization. Let the input of MambaTAD be videos $\mathbf{X}$, consisting of a series of frames or video clips features extracted at different time steps $t$ by pre-trained models~\cite{carreira2017quo, alwassel2021tsp, tong2022videomae, wang2023videomae, wang2024internvideo2}, represented as $ \{\mathbf{x}_1, \mathbf{x}_2, \ldots, \mathbf{x}_t, \ldots, \mathbf{x}_T\}$, where the total duration $T$ is different across videos. For simplicity, $x_t$ and $\mathbf{x}_t$ both denote the $t$-th discrete token, where $\mathbf{x}_t\in\mathbb{R}^c$ in video feature representation. MambaTAD learns to model the underlying discrete signal resulting from videos and the action label $\mathbf{Y} = \{\mathbf{y}_1, \mathbf{y}_2, \ldots, \mathbf{y}_I\}$, where $I$ is the number of ground truth labels in a video and $\mathbf{y}_i = (start_i, end_i, action_i)$ to produce action proposals $(start_i, end_i)$ and categories $action_i$ during inference. Here, $start_i, end_i\in[0,T]$ and $start_i<end_i$, $action_i\in\{0, C-1\}$ where $C$ is pre-defined categories. 

\begin{figure}[t]
    \centering
    \includegraphics[width=1.0\linewidth]{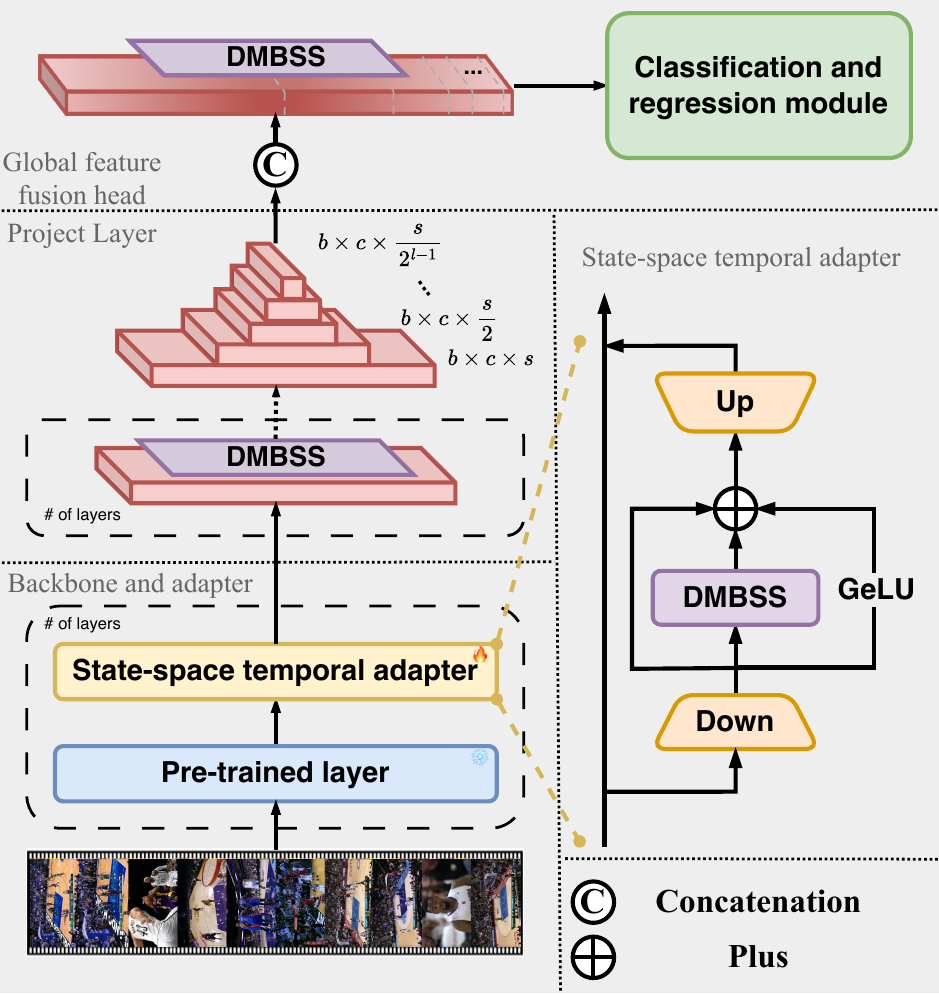}
    \caption{The overall MambaTAD architecture uses large-scale pre-trained models for the backbone, with a State-Space Temporal Adapter (SSTA) in the end-to-end setting. Pyramid features are processed by Diagonal-Masked Bidirectional State-Space (DMBSS) modules, followed by a global fusion head that progressively concatenates features for global context.}
    \label{fig:mambatad}
\end{figure}
\begin{figure*}[t]
\centering
\includegraphics[width=0.8\textwidth]{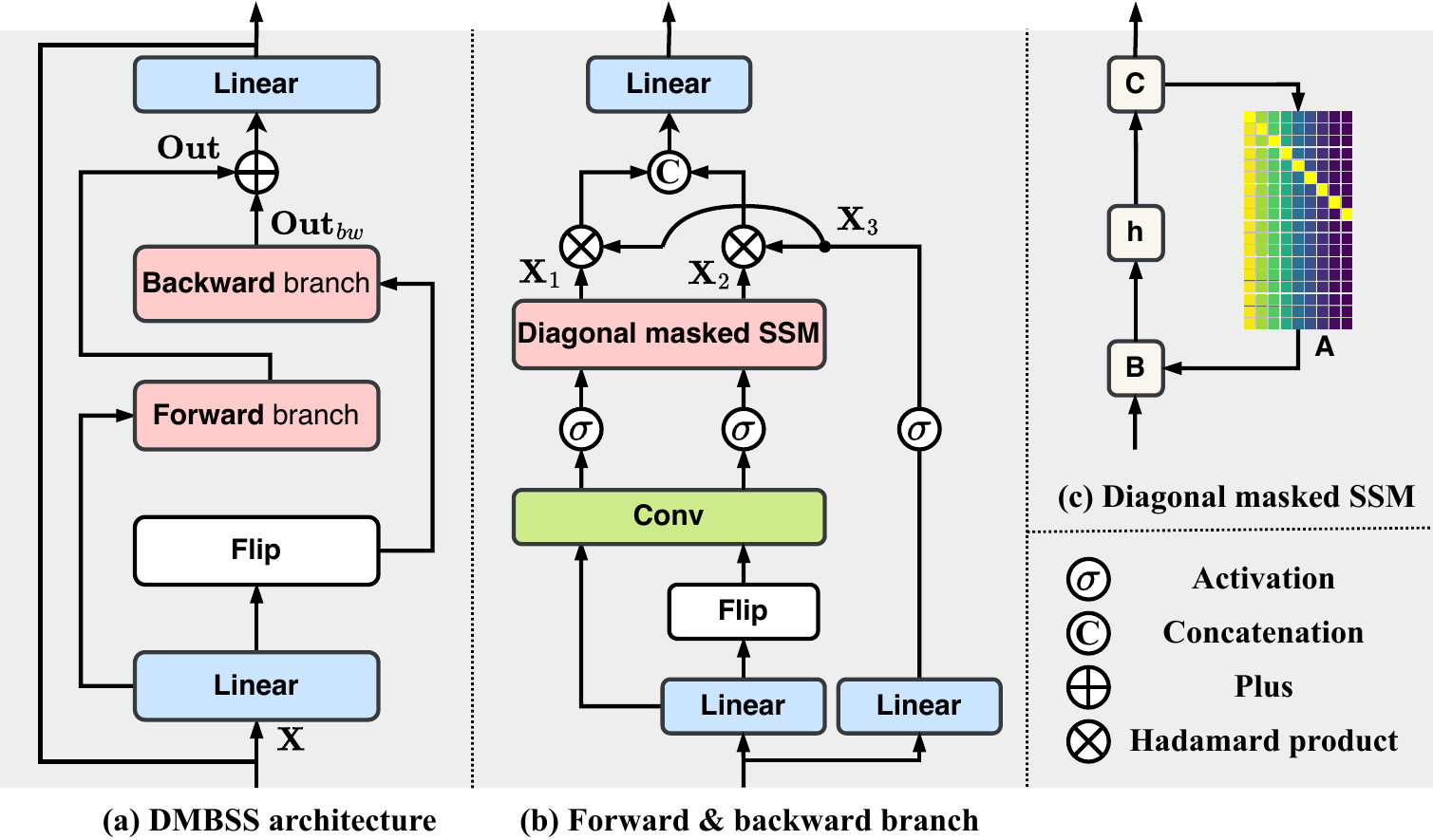} 
\caption{(a) The architecture of DMBSS. (b) Forward and backward branches share the same architecture. (c) We mask the diagonal elements in the state transformation matrix $A$ to solve the self-elements conflict. }
\label{archi}
\end{figure*}

\subsection{The Framework}
The framework applies a Parameter-efficient fine-tuning (PEFT) mechanism to fine-tune a plug-and-play module named State-Space Temporal Adapter (SSTA) to achieve efficient and effective transfer learning for end-to-end temporal action detection and build a simple yet effective anchor-free action detector. As shown in Fig. \ref{fig:mambatad}, MambaTAD consists of three main parts: a video backbone, a projection pyramid with Diagonal-masked Bidirectional State-Space (DMBSS) modules, and a global feature fusion head. We begin by leveraging video backbones \cite{carreira2017quo, alwassel2021tsp, tong2022videomae, wang2023videomae, wang2024internvideo2} to extract spatiotemporal features. To efficiently model long-range dependencies in an end-to-end manner, we introduce SSTA that captures both local and global non-causal information with a smaller parameter overhead.
It is worth noting that we consider both end-to-end (E2E) and non-end-to-end settings, while SSTA is utilized exclusively in the end-to-end setting. Following that, we construct pyramid projection layers incorporating the proposed DMBSS to model action instances across varying temporal durations. While recent TAD methods \cite{zhang2022actionformer, shi2023tridet, gan2022temporal} iteratively downsample temporal features, our approach uniquely applies DMBSS at each scale to enhance cross-scale feature interactions. Lastly, we fuse multi-granularity representations from different pyramid levels to improve global contextual understanding, leading to more precise action detection. We will elaborate on the proposed modules as follows.

\subsection{Diagonal-Masked Bidirectional State-Space Module} \label{sec:DMBSS}





Given an input $\mathbf{X}\in \mathbb{R}^{s \times c}$ where $s$ denotes the sequence length after feature sampling, proportional to the original video duration $T$, and $c$ is the feature channel, the output $\mathbf{X}_{out}$ maintains the same dimensional structure as the input. Mamba \cite{gu2023mamba} is originally designed for the 1D sequence under causal relations in language tasks. Following \cite{dao2024transformers}, 
the final output in Eq. \ref{eqn-4} can be simplified as:
\begin{equation}\small
    \mathbf{Y} = \mathcal{C}\cdot(\mathbf{M\cdot X}), 
    \label{eq6}
\end{equation}
where $\mathcal{C}$ is a diagonal matrix which can be neglected in derivation and $\mathbf{M}$ is a lower triangle matrix:
\begin{equation}\small
    \mathcal{C}=\text{diag}({\mathbf{C}}_1,{\mathbf{C}}_2, \cdots, {\mathbf{C}}_s)
    \label{eq7}
\end{equation}
\begin{equation}\small
\!\!\!\!\mathbf{M}\!=\!\!\begin{bmatrix}
    \bar{\mathbf{B}}_1 & 0 & \!\cdots\! & 0 \\
    \bar{\mathbf{A}}_2\bar{\mathbf{B}}_1 & \bar{\mathbf{B}}_2 & \!\cdots\! & 0  \\
    \vdots & \vdots & \!\ddots\! & \vdots \\
    \!\bar{\mathbf{A}}_s\bar{\mathbf{A}}_{s-1}\!\cdots\!\bar{\mathbf{A}}_2\bar{\mathbf{B}}_1\!\! &  \!\bar{\mathbf{A}}_s\bar{\mathbf{A}}_{s-1}\!\cdots\! \bar{\mathbf{A}}_3\bar{\mathbf{B}}_2\! & \!\cdots\! & \!\bar{\mathbf{B}}_s\!\!
    \end{bmatrix}.
\end{equation}
From $\mathbf{M}$, it is evident that the earlier state tends to exhibit more zero weights, a phenomenon referred to as causality. In NLP and text generation, causality is fundamental \cite{hu2021causal}, as autoregressive models generate text sequentially, relying solely on preceding tokens to maintain logical coherence and prevent future information from influencing past decisions. However, in the TAD task, a video sequence’s future state depends not only on prior tokens but also on the broader temporal context, where identifying action boundaries requires information from both preceding and succeeding tokens. As a result, enforcing strict causality leads to temporal information loss by restricting the model’s ability to incorporate future context for more precise action localization. On the other hand, bidirectional temporal information plays a key role in video representation modeling. \cite{zhu2017bidirectional} exploits bidirectional processing in combination with multirate feature reconstruction to integrate both past and future context across multiple temporal resolutions, effectively mitigating information loss over long sequences. Similarly, a Decomposed Bidirectionally Mamba block (DBM)\cite{chen2024video} applies a Mamba \cite{gu2023mamba} module with shared parameters for bidirectional scanning, allowing more effective temporal feature processing, which partially mitigates this issue. Specifically, let the feature sequence of length $s$ with channel dimension $c$ be concatenated as:
\begin{equation}\small
    \mathbf{X}_{fw} = \begin{bmatrix}
        \mathbf{x}_1^T &\mathbf{x}_2^T & \cdots &\mathbf{x}_s^T
    \end{bmatrix}^T \in \mathbb{R}^{s \times c},
\end{equation}
For a clear and rigorous theoretical exposition, we define the reversal matrix
\begin{equation} \small
    \mathbf{J}_s\in\{0,1\}^{s\times s},\qquad (\mathbf{J}_s)_{i,j}=
\begin{cases}
1,& j=s+1-i,\\
0,& \text{otherwise},
\end{cases}
\end{equation}
such that the backward sequence is given by
\begin{equation} \small
    \mathbf{X}_{bw}=\mathbf{J}_s \mathbf{X}_{fw}.
\end{equation}
Accordingly, we obtain
\begin{equation} \small
\begin{aligned}
    \mathbf{Y}_{fw} = \mathbf{M}_{fw} \mathbf{X}_{fw} , 
    \mathbf{Y}_{bw} = \mathbf{M}_{bw} \mathbf{X}_{bw}, 
    \mathbf{Y} = \mathbf{Y}_{fw} + \mathbf{J}_s \mathbf{Y}_{bw}.
\end{aligned}
\end{equation}
The matrices $\mathbf{M}_{fw}$ and $\mathbf{M}_{bw}$ undergo a transformation wherein their constituent elements $\bar{\mathbf{A}}_i\cdots\bar{\mathbf{B}}_j$ are replaced by the codes $fw_{ij}$ and $bw_{ij}$, respectively. Consequently, the forward and backward representations can be written as
\begin{equation}
    \mathbf{Y}_{fw} = \begin{bmatrix}
        fw_{11} & 0 & \cdots &0 \\
        fw_{21} & fw_{22} & \cdots & 0 \\
        \vdots & \vdots & \ddots & \vdots \\
        fw_{s1} & fw_{s2} & \cdots &fw_{ss}
    \end{bmatrix}
    \begin{bmatrix}
        \mathbf{x}_1^T \\
        \mathbf{x}_2^T \\
        \vdots \\
        \mathbf{x}_s^T 
    \end{bmatrix}
    = \begin{bmatrix}
        h_{1f} \\
        h_{2f} \\
        \vdots \\
        h_{sf}
    \end{bmatrix}
\end{equation}
and
\begin{equation}
\begin{aligned}
    \mathbf{Y}_{bw}\!\!=\!\! \begin{bmatrix}
        bw_{ss} & 0 & \cdots & 0 \\
        bw_{(s-1)s}\!\! & bw_{(s-1)(s-1)}\!\! & \cdots & 0 \\
        \vdots & \vdots & \ddots & \vdots \\
        bw_{1s} & bw_{1(s-1)} & \cdots & bw_{11}
    \end{bmatrix}\!\!\!\!
    \begin{bmatrix}
        \mathbf{x}_s^T \\
        \!\mathbf{x}_{s-1}^T\!\! \\
        \vdots \\
        \mathbf{x}_1^T 
    \end{bmatrix} 
    \!\!=\!\! \begin{bmatrix}
        h_{sb} \\ \!h_{(s-1)b}\!\! \\ \vdots \\ h_{1b}
    \end{bmatrix}\!\!.
\end{aligned}
\end{equation}
By combining the forward and backward components, each hidden state can then be expressed as
\begin{equation}
\!\!\!\begin{aligned}
    h_{1} &= h_{1f}+h_{1b} 
    = fw_{11}\mathbf{x}_1^T + bw_{1s}\mathbf{x}_s^T + bw_{1(s-1)}\mathbf{x}_{s-1}^T \\
    &+ \cdots + bw_{11}\mathbf{x}_1^T, \\
    h_2 &= h_{2f} + h_{2b} 
    = fw_{21}\mathbf{x}_1^T + fw_{22}\mathbf{x}_2^T + bw_{2s}\mathbf{x}_s^T \\
    &+ bw_{2(s-1)}\mathbf{x}_{s-1}^T + \cdots + bw_{22}\mathbf{x}_2^T, \\
    h_s &= h_{sf} + h_{sb} = fw_{s1}\mathbf{x}_1^T + fw_{s2}\mathbf{x}_2^T + \cdots + fw_{ss}\mathbf{x}_s^T \\
    &+ bw_{ss}\mathbf{x}_s^T .
\end{aligned}
\end{equation}
This derivation yields the final representation:
\begin{equation}  \small
\begin{aligned}
\mathbf{Y} \!\!=\!\! \begin{bmatrix}
    h_1 \\ h_2 \\ \vdots \\ h_s \\
\end{bmatrix}
\!\!&=\!\! \begin{bmatrix}
        fw_{11}\!+\!bw_{11} & bw_{12} & \cdots & bw_{1s} \\
        fw_{21} & fw_{22}\!+\!bw_{22} & \cdots & bw_{2s} \\
        \vdots & \vdots & \ddots & \vdots \\
        fw_{s1} & fw_{s2} & \cdots & fw_{ss}\!+\!bw_{ss}
    \end{bmatrix}\!\!\!\!
    \begin{bmatrix}
\mathbf{x}_1^T \\
\mathbf{x}_2^T \\
\vdots \\
\mathbf{x}_s^T
\end{bmatrix} \\
&= \mathbf{M}' \mathbf{X}.
\end{aligned}
    \label{eq11}
\end{equation}

We can see the final weights matrix $\mathbf{M}^{\prime}$ before $\mathbf{X}$ contains redundant representations of the self-similarity of each token. As a result, the self-similarity is effectively doubled within these diagonal elements, potentially weakening the associations with other tokens. In the context of TAD, the aim is to elucidate connections between distinct elements, \textit{viz.}, start and end positions; therefore, efforts should be made to mitigate the duplication of self-similarity present in the diagonal and increase feature discrimination. To achieve this, we set the diagonal elements in each backward pass to 0, removing redundancy and improving inter-token relationship clarity. 

As illustrated in Alg. \ref{alg:algorithm}, we expand the feature dimensionality to $\lambda c$ (with $\lambda=4$) via a linear transformation after LayerNorm, enabling dual-branch bidirectional processing. The expanded representation is split into two equal parts for forward and backward streams. To enforce bidirectional parameter sharing \cite{chen2024video}, the second segment is inverted before concatenation. Besides, we propose a dual-branch approach, without parameter sharing, that enhances temporal modeling and mitigates the decay of temporal information via two separate branches. Our ablation study in Sec. \ref{sec:ablation_dmbss} gives experimental analysis for this. To mitigate the issue of diagonal conflict, we introduce a masking mechanism that nullifies the diagonal elements in the trainable weights for each backward branch. The $SSM(\cdot)$ in Alg. \ref{alg:algorithm} stands for vanilla mamba \cite{gu2023mamba} without output linear layer, thereby reducing the output dimensionality from $2c$ to $c$. That is because the application of a final linear transformation at the end of the dual-branch processing enhances dual-branch feature fusion. 
After diagonal masking, bidirectional dual branches integrate to capture richer temporal contexts. The representations are split along the batch dimension and then merged along the channel dimension for consolidation. Lastly, a final linear layer integrates both temporal and channel information from two separate branches and restores the original channel size, yielding the output token. 

\begin{algorithm}[t]\small
\caption{Dual branch DMBSS.}
\label{alg:algorithm}
\textbf{Require}: token sequence $\mathbf{X}_{m-1}$: \textcolor{gray}{$(b,s,c)$} \\
\textbf{Ensure}: token sequence $\mathbf{X}_m$: \textcolor{gray}{$(b,s,c)$} 
\begin{algorithmic}[1]  
\Function{$\mathbf{mask\_diagonal}$}{$\mathbf{Mat}$}
    \State $\mathbf{row}, \mathbf{col} \gets \mathbf{Mat}.\mathrm{size}(0), \mathbf{Mat}.\mathrm{size}(1)$
    \State $\mathbf{diagonal\_idx} \gets \mathrm{range}(\min(\mathbf{row}, \mathbf{col}))$
    \State $\mathbf{Mat}[\mathbf{diagonal\_idx},\mathbf{diagonal\_idx}] \gets 0$
    \State \Return $\mathbf{Mat}$
\EndFunction
\State $\mathbf{x}$: \textcolor{gray}{$(b,4c,s)$} $\gets \mathbf{Linear}^{\mathbf{x}}(\mathbf{X}_{m-1})$
\State $\mathbf{x}_{fw}, \mathbf{x}_{bw}$: \textcolor{gray}{$(b,2c,s)$} $\gets \mathbf{Chunk}(\mathbf{x})$
\State $\mathbf{x}$: \textcolor{gray}{$(2b,2c,s)$} $\gets [\mathbf{x}_{fw}, \mathbf{x}_{bw}.\mathrm{flip}]$ 
\State $\mathbf{A}$: \textcolor{gray}{$(c,d)$} $\gets -\exp(\mathbf{Parameter}_o^\mathbf{A})$    
\State $\mathbf{A}_{fw}, \mathbf{A}_{bw}$: \textcolor{gray}{$(\frac{c}{2},d)$} $\gets \mathbf{Chunk}(\mathbf{A})$
\State $\mathbf{A}$: \textcolor{gray}{$(c,d)$} $\gets [\mathbf{A}_{fw}, \mathbf{mask\_diagonal}(\mathbf{A}_{bw})]$  
\State $\mathbf{A}_{bw}$: \textcolor{gray}{$(c,d)$} $\gets -\exp(\mathbf{Parameter}_o^{\mathbf{mask\_diagonal}(\mathbf{A}^\prime)})$
\State $\mathbf{Z}$: \textcolor{gray}{$(2b,c,s)$} $\gets \mathbf{SSM}(\mathbf{x}, \mathbf{A})$
\State $\mathbf{Z}_{bw}$: \textcolor{gray}{$(2b,c,s)$} $\gets \mathbf{SSM}(\mathbf{x}.\mathrm{flip}, \mathbf{A}_{bw})$
\State $\mathbf{Z}_{fw}, \mathbf{Z}_{bw}$: \textcolor{gray}{$(b,c,s)$} $\gets \mathbf{Chunk}(\mathbf{Z} + \mathbf{Z}_{bw}.\mathrm{flip})$
\State $\mathbf{out}$: \textcolor{gray}{$(b,2c,s)$} $\gets [\mathbf{Z}_{fw}, \mathbf{Z}_{bw}.\mathrm{flip}]$
\State $\mathbf{X}_m$: \textcolor{gray}{$(b,s,c)$} $\gets \mathbf{Linear}^{\mathbf{Z}}(\mathbf{Z})$
\State \Return $\mathbf{X}_m$
\end{algorithmic}
\end{algorithm}

\noindent\textbf{Implementation.} 
The input feature $\mathbf{X}$ will go through two parallel branches for better temporal context representation as illustrated in Fig. \ref{archi}. Then there are three distinct pathways for processing sequences in each branch. The first two pathways integrate information in both the forward and backward directions with shared parameters, while the third pathway processes information independently:
\begin{equation}  \small\small
\begin{aligned}
    \mathbf{X}_1&=SSM(\sigma(DWConv(Linear(\mathbf{X})))), \\
    \mathbf{X}_2&=SSM(\sigma(DWConv(Linear(flip(\mathbf{X}))))), \\
    \mathbf{X}_3&=\sigma(Linear(\mathbf{X})),
\end{aligned}
\end{equation}
where $flip(\cdot)$ is flipping operation along the sequence dimension, $DWConv(\cdot)$ represents depth-wise convolution, $SSM(\cdot)$ is selective scan mechanism ~\cite{gu2023mamba} without out linear layer and $\sigma(\cdot)$ denotes SiLU activation~\cite{shazeer2020glu}. Following ~\cite{chen2024video}, features from the three pathways are aggregated with element-wise product and concatenation:
\begin{equation}  \small\small
\mathbf{out}=(\mathbf{X}_1\odot\mathbf{X}_3)\oplus(\mathbf{X}_2\odot\mathbf{X}_3).
\end{equation}
In the backward branch, the overall structure parallels that of the forward branch, with the distinction that the input is the flip of $\mathbf{X}$, and the output is denoted as $\mathbf{out}_{bw}$.  To enhance dual-branch feature fusion, features from both branches are combined and projected back to dimension $c$ to produce the output $\mathbf{X}_{out}$, which retains the same shape as the input and captures richer temporal context. Finally, a residual connection is used to facilitate gradient flow and improve training stability. The output can be represented as:
\begin{equation}  \small
    \mathbf{X}_{out}=Linear(\mathbf{out}+\mathbf{out}_{bw}) + \mathbf{X}.
\end{equation}



\subsection{Building Projection Layers}\label{sec:proj}
The blocks designed in previous TAD networks \cite{zhang2022actionformer, shi2023tridet, yang2024dyfadet} mainly follow the $Norm\rightarrow Attention/CNN \rightarrow Norm \rightarrow MLP$ flow in each pyramid layer. 
To address challenges related to computational complexity \cite{DBLP:conf/icml/ZhuL0W0W24}, feature discrimination \cite{shi2023tridet, yang2024dyfadet}, and receptive field limitations \cite{luo2016understanding, guo2024mambair}, we employ Mamba \cite{gu2023mamba} to model global dependencies. However, in Section \ref{sec:DMBSS}, we analyze that the vanilla Mamba's causality mechanism leads to a loss of temporal information, making it unsuitable for the TAD task. This underscores the potential of the proposed DMBSS block in effectively capturing action features for improved localization. The experimental results in Table \ref{table2} in Section IV-F further validate our analysis.

\noindent\textbf{The DMBSS projection layer.}
In this section, we construct projection layers utilizing the DMBSS discussed previously. As shown in Fig. \ref{fig:mambatad}, each projection layer consists of a DMBSS and a Max pooling layer \cite{tang2023temporalmaxer}, because we aim to preserve the benefits of long sequence modeling while enhancing the discriminability of temporal features, all within the linear computational complexity. The two blocks are given by:
\begin{equation}\small
    \begin{aligned}
        \mathbf{f}_l ^{\prime}&=DMBSS(LN(\mathbf{f}_{l-1})) + \mathbf{f}_{l-1}, \\
        \mathbf{f}_l&=MaxPool(\mathbf{f}_l ^{\prime}),
    \end{aligned}
\end{equation}
where $DMBSS(\cdot)$ is the block proposed previously, and $MaxPool(\cdot)$ is the max pooling for temporal feature. 

The feature pyramid derived from the state-space model outperforms earlier transformer-based and CNN-based feature pyramids, delivering enhanced performance while also ensuring greater computational efficiency. 
DMBSS's rationale and efficiency for enhancing TAD tasks are deeply rooted in leveraging the strengths of the S6 model and our improvements. More specifically, it retains the recursive structure outlined in Eq. (3), which empowers the model to retain information from exceptionally long sequences without temporal information decay and self-conflict, allowing for a greater number of features to contribute to the detection process and more precise feature discrimination. Concurrently, the incorporation of a dual-branch parallel scan algorithm allows MambaTAD to achieve the same computational efficiencies as that in Eq. \ref{eqn-4}, ensuring computational effectiveness.

\subsection{Global Feature Fusion Head}\label{sec:head}
The detection head plays a pivotal role in TAD, as it is directly linked to the final prediction outcome. 
Inspired by \cite{yu2024mambaout, zhu2024temporally}, it is evident that the visual state-space model excels in tasks characterized by long sequences. Consequently, we innovatively concatenate features from different levels of the pyramid sequentially to create an extended sequence:
\begin{equation}  \small
\mathbf{F}=\mathbf{f}_1\oplus\mathbf{f}_2\oplus\ldots\oplus\mathbf{f}_l,
\end{equation}
where $\mathbf{F}\in\mathbb{R}^{b\times c\times (s+{s\over 2}+\ldots+{s\over 2^{l-1}})}$, and $l$ is the layer of pyramid. To address the lack of global awareness caused by hierarchical feature structure~\cite{lin2017feature, liu2018path}, a global feature fusion mechanism is designed to ensure the full use of long-range information across different layers and improve regression and classification accuracy. To this end, we propose the residual global level detection head to adapt DMBSS for action detection. As shown in Fig. \ref{archi}, we first use the LayerNorm (LN) followed by the DMBSS to capture the temporal long-range dependencies. Then, a residual connection is utilized for easier optimization. Lastly, we use a successive LN to ensure stable feature distribution. The final output can be mathematically represented as follows:
\begin{equation}  \small
    \mathbf{F}_G = LN(DMBSS(LN(\mathbf{F}))+\mathbf{F}).
\end{equation}

By concatenating features from different pyramid levels into an extended sequence, our design enables the detection head to access temporal information across multiple granularities simultaneously. Traditional hierarchical feature pyramid approaches~\cite{lin2017feature, zhang2022actionformer} process each level independently, which limits the receptive field and reduces global context. In contrast, the extended sequence allows DMBSS to operate over a longer, contiguous feature stream, capturing temporal long-range dependencies across all pyramid layers. The residual global feature fusion further stabilizes the feature distribution and facilitates optimization. Together, this design enhances global awareness and improves both classification and boundary regression accuracy in temporal action detection.

\subsection{State-Space Temporal Adapter}
In Sections \ref{sec:proj} and \ref{sec:head}, we construct the action detector with projection layers and a global feature fusion head. End-to-end TAD demands strong temporal modeling to bridge low-level encoding and high-level action detection. However, fine-tuning the entire backbone is costly \cite{liu2024end} and may hinder generalization. On the other hand, it remains unexplored whether Mamba can serve as an adapter to bridge large pre-trained foundational models\cite{tong2022videomae,wang2023videomae, wang2024internvideo2} and downstream video understanding tasks. To address these, we propose SSTA, a lightweight yet expressive mechanism for refining temporal representations efficiently.

The standard adapter \cite{houlsby2019parameter} is widely used in PEFT but focuses on the channel dimension, limiting its ability to capture temporal dependencies. AdaTAD \cite{liu2024end} addresses this with the Temporal-Informative Adapter (TIA), combining temporal depth-wise convolution and a fully connected layer. However, depth-wise convolution constrains the receptive field \cite{luo2016understanding, ding2022scaling}, while the fully connected layer increases trainable parameters.
To overcome these challenges, we propose the State-Space Temporal Adapter (SSTA), as illustrated in Fig. \ref{fig:mambatad}. SSTA integrates the DMBSS to enhance temporal modeling. This design enables bidirectional feature interaction while maintaining controlled diagonal masking, effectively capturing both long-range dependencies and local discriminative patterns. The formulation of SSTA is given by:
\begin{equation}\small
    \begin{aligned}
        \hat{\mathbf{x}} &= \mathbf{W}^T_{down}\mathbf{x}, \\
        \bar{\mathbf{x}} &=\theta(\hat{\mathbf{x}}), \\
        {\mathbf{x}}^{\prime} &= DMBSS(\hat{\mathbf{x}})+\hat{\mathbf{x}} + \bar{\mathbf{x}}, \\
        {\mathbf{x}}^{\prime\prime} &= \mathbf{W}^T_{up} \mathbf{x}^{\prime} + \mathbf{x}, \\
    \end{aligned}
\end{equation}
where $\mathbf{x}, \mathbf{x}^{\prime\prime} \in \mathbb{R}^{t \times h \times w \times d}$ denote input and output features, $t$ represents the number of frames, $h$ and $w$ correspond to the spatial dimensions of the video, and $d$ is the hidden feature dimensionality. The function $DMBSS(\cdot)$ represents our proposed block. Following AdaTAD\cite{liu2024end}, we employ GELU \cite{hendrycks2016gaussian} as the non-linear activation function, denoted as $\theta(\cdot)$. The transformation matrices $\mathbf{W}^T_{down} \in \mathbb{R}^{d \times \frac{d}{\lambda}}$ and $\mathbf{W}^T_{up} \in \mathbb{R}^{\frac{d}{\lambda} \times d}$ are trainable projection weights, where $\lambda$ is a scalable factor which controls intermediate dimensions. $\lambda > 1$ is enforced to ensure parameter efficiency and reduction. One residual connection is used for better convergence.

By embedding SSTA within our end-to-end TAD pipeline, we achieve global context refinement while maintaining a computationally efficient framework. The DMBSS-driven state-space transition ensures that both local short-term dynamics and long-term action dependencies are effectively captured. Our experiments show that SSTA can achieve better performance with less memory usage and computations.

\begin{table*}[ht]
\centering
\caption{Comparison of different TAD methods on THUMOS14 and ActivityNet-1.3 with off-the-shelf features. The performance is evaluated using mAP at different IoU thresholds. $\dagger$ stands for the feature fused with the optical flow; * indicates the results of our re-implementations. 
For ActivityNet-1.3, the average mAP is calculated over [0.5:0.95:0.05].
} 
\vspace{-1mm}
\setlength{\tabcolsep}{7.0pt}
\begin{tabular}{l||c|c|c|c|c|c|c|c||c|c|c|c|c}
\toprule
\multirow{2}{*}{Method} &\multirow{2}{*}{E2E}& \multicolumn{7}{c||}{THUMOS14} & \multicolumn{5}{c}{ActivityNet-1.3} \\
\cmidrule{3-14}
&&  Backbone & 0.3 & 0.4 & 0.5 & 0.6 & 0.7 & Avg. & Backbone & 0.5 & 0.75 & 0.95 & Avg. \\
\midrule
BMN \cite{lin2019bmn} & $\usym{2717}$ &  TSN$\dagger$ & 56.0 & 47.4 & 38.8 & 29.7 & 20.5 & 38.5 & TSN$\dagger$ & 50.1 & 34.8 & 8.3 & 33.9 \\
PCG-TAL \cite{su2020pcg} & $\usym{2717}$&  I3D$\dagger$ & 65.1 & 59.5 & 51.2 & - & - & - & I3D$\dagger$ & 52.0 & 35.9 & 8.0 & 34.9 \\
TCANet \cite{qing2021temporal} & $\usym{2717}$&   TSN$\dagger$ & 60.6 & 53.2 & 44.6 & 36.8 & 26.7 & 44.3 & SlowFast & 54.3 & 39.1 & 8.4 & 37.6 \\
RTD-Net \cite{tan2021relaxed} & $\usym{2717}$&  I3D$\dagger$ & 68.3 & 62.3 & 51.9 & 38.8 & 23.7 & 49.0 & I3D$\dagger$ & 47.2 & 30.7 & 8.6 & 30.8 \\
MCBD \cite{su2023multi} & $\usym{2717}$&  TSN$\dagger$ & 68.38 & 62.2 & 53.5 & 41.6 & 29.5 & 51.1 & R(2+1)D & 49.2 & 34.2 & 10.1 & 33.6 \\
VSGN \cite{zhao2021video} & $\usym{2717}$&   TSN$\dagger$ & 66.7 & 60.4 & 52.4 & 41.0 & 30.4 & 56.6 & TSN$\dagger$ & 52.4& 36.0& 8.4& 35.1 \\
TadTR \cite{liu2022end} & $\usym{2717}$&   I3D$\dagger$ & 74.8 & 69.1 & 60.1 & 46.6 & 32.8 & 56.7 & TSN$\dagger$ & 53.6 & 37.5 & 10.6 & 36.8 \\
ActionFormer \cite{zhang2022actionformer} & $\usym{2717}$ & I3D$\dagger$ & 82.1 & 77.8 & 71.0 & 59.4 & 43.9 & 66.8 & R(2+1)D & 54.7 & 37.8 & 8.4 & 36.6 \\
DiffTAD \cite{nag2023difftad} & $\usym{2717}$&  I3D$\dagger$ &74.9 &72.8 &71.2& 62.9& 58.5& 68.0&I3D$\dagger$ & 56.1& 36.9& 9.0& 36.1 \\
TriDet \cite{shi2023tridet} & $\usym{2717}$ &  I3D$\dagger$ & 83.6 & 80.1 & 72.9 & 62.4 & 47.4 & 69.3 & SlowFast & 54.7 & 38.0 & 8.4 & 36.8 \\
DyFADet \cite{yang2024dyfadet} & $\usym{2717}$ &  I3D$\dagger$ & 84.0 &80.1 &72.7 &61.1 &47.9 &69.2 & R(2+1)D &58.1 &39.6 &8.4 &38.5 \\
LFAF \cite{tang2024learnable} & $\usym{2717}$ &  I3D$\dagger$ & 83.0 &79.5 &73.8 &62.5 &\textbf{48.2} &69.4 & I3D$\dagger$ &59.3 &39.8 &7.8 &39.0 \\
\rowcolor{gray!25} \textbf{Ours} & $\usym{2717}$&  I3D$\dagger$ & \textbf{84.3} & \textbf{80.7} & \textbf{74.1} & \textbf{62.9} & 47.5 & \textbf{69.9} & R(2+1)D & \textbf{60.2} & \textbf{41.3} & \textbf{9.7} & \textbf{40.2} \\
\midrule
VideoMAE v2 \cite{wang2023videomae} & $\usym{2717}$ & VM2-g & 84.0 &79.6 &73.0 &63.5 &47.7 &69.6 & -&-&-&-&-\\
DyFADet \cite{yang2024dyfadet} & $\usym{2717}$ &  VM2-g$\dagger$ & 85.4 & - & - & - & 50.2 & 71.1 & - &- &- &- &- \\
ADSFormer \cite{10440542} & $\usym{2717}$  & VM2 & 85.3 & 80.8 & 73.9 & 64.0& 49.8 & 70.8 & VM2 & 55.3 & 38.4 & 8.4 & 37.1 \\
InternVideo2*~\cite{wang2024internvideo2} & $\usym{2717}$& IV6B &86.9 & 81.9& 75.1 & 65.8 & 50.3&72.0& IV6B &61.5 & \textbf{44.6} & \textbf{12.7}& 41.2 \\ 
VideoMambaSuite~\cite{chen2024video} & $\usym{2717}$&IV6B& 86.9 &83.1& 76.9& 65.9& 50.8& 72.7& IV6B& 62.4& 43.5& 10.2& 42.0 \\
\rowcolor{gray!25} \textbf{Our MambaTAD} & $\usym{2717}$&  IV6B & \textbf{87.5} & \textbf{83.8} & \textbf{78.3} & \textbf{67.3} & \textbf{52.9} & \textbf{73.9} & IV6B & \textbf{63.1} & 44.2 & 11.0 & \textbf{42.8} \\
\bottomrule
\end{tabular}
\label{tab:stand_comparison}
    \vspace{-1mm}
\end{table*}

\section{Experiments}
\subsection{Experimental Setup}
\noindent\textbf{Model Setting. } We explore both non-end-to-end and end-to-end settings—one leveraging off-the-shelf features and the other using the videos as inputs. SSTA is exclusively employed in the end-to-end setting to fine-tune the feature extractor. Please refer to \textit{Supp.} \ref{app:imp} for implementation details.

\noindent\textbf{Dataset.}
We experiment on five challenging datasets: THUMOS14~\cite{thumos14}, ActivityNet-1.3~\cite{caba2015activitynet}, MultiThumos\cite{yeung2018every}, HACS-Segment~\cite{zhao2019hacs}, and FineAction~\cite{liu2022fineaction}. 
The THUMOS14 dataset includes 20 sports action categories, with 200 training and 213 testing untrimmed videos, covering 3,007 and 3,358 action instances. ActivityNet-1.3 features 200 categories across approximately 20K videos and 600 hours, divided into 10K training, 4,926 validation, and 5,044 test videos. MultiThumos introduces more complex scenarios, presenting a greater challenge for TAD. It extends the THUMOS14 dataset with dense, multilabel, frame-level action annotations spanning 30 hours across 400 videos. The dataset comprises 38,690 annotations covering 65 action classes, with an average of 1.5 labels per frame and 10.5 action classes per video, significantly increasing the task's difficulty. 
HACS-Segment has 200 categories, with 37.6K training and 5,981 test videos. FineAction offers 103K temporal instances across 106 fine-grained actions in 17K untrimmed videos, with 8,440 training, 4,174 validation, and 4,118 testing. Only training and validation annotations are available. We adhere to standard protocols for training and evaluation.

\noindent\textbf{Evaluation metrics. }
We employ a typical evaluation metric used in the TAD task: the standard mean Average Precision (mAP) at various Intersection over Union (IoU) thresholds. 

\begin{table*}[ht]
\centering
\caption{Comparison of different TAD methods on THUMOS14 and ActivityNet-1.3 under end-to-end setting. Mem refers to memory usage per video.
For ActivityNet-1.3, the average mAP is calculated over [0.5:0.95:0.05].
}
\setlength{\tabcolsep}{5.0pt}
\begin{tabular}{l||c|c|c|c|c|c|c|c|c||c|c|c|c|c}
\toprule
\multirow{2}{*}{Method} &\multirow{2}{*}{E2E}&\multirow{2}{*}{Mem}& \multicolumn{7}{c||}{THUMOS14} & \multicolumn{5}{c}{ActivityNet-1.3} \\
\cmidrule{4-15}
&& & Backbone & 0.3 & 0.4 & 0.5 & 0.6 & 0.7 & Avg. & Backbone & 0.5 & 0.75 & 0.95 & Avg. \\
\midrule
AFSD \cite{lin2021learning} & $\usym{2713}$& 12G&  I3D$\dagger$ & 67.3 & 62.4 & 55.5 & 43.7 & 31.1 & 52.0 & I3D$\dagger$ & 52.4 & 35.3 & 6.5 & 34.4 \\
E2E-TAD \cite{liu2022empirical} & $\usym{2713}$ &12G& SlowFast & 69.4& 64.3 &56.0 &46.4 &34.9 &54.2 & SlowFast & 50.5& 36.0& 10.3& 35.1\\
BasicTAD \cite{yang2023basictad} & $\usym{2713}$ &12G& SlowOnly & 75.5 &70.8 &63.5 &50.9& 37.4& 59.6 & SlowOnly & 51.2& 33.4& 7.6& 33.1\\
TALLFormer \cite{cheng2022tallformer} & $\usym{2713}$& 29G & Swin & 76.0 & - & 63.2 & - & 34.5 & 59.2 & Swin & 54.1 & 36.2 & 7.9 & 35.6 \\
Re$^2$TAL \cite{zhao2023re2tal} & $\usym{2713}$& 24G& Re$^2$VideoSwin-T&77.0& 71.5& 62.4& 49.7& 36.3& 59.4& Re$^2$VideoSwin-T& 54.8& 37.8&9.0& 36.8 \\
ViT-TAD \cite{yang2024adapting} & $\usym{2713}$& -&ViT-B & 85.1 &80.9 &74.2 &61.8 &45.4 &69.5 & ViT-B & 55.9& 38.5& 8.8& 37.4 \\
AdaTAD \cite{liu2024end} &$\usym{2713}$ & 2.5G & VideoMAE-S & 84.5 &80.2& 71.6& 60.9& 46.9& 68.8& VideoMAE-S& 56.2 &39.0 &9.1 &37.9 \\
AdaTAD \cite{liu2024end} &$\usym{2713}$ & 4.9G &VideoMAE-B& 87.0& 82.4& 75.3& 63.8& 49.2& 71.5& VideoMAE-B & 56.8 &39.4& 9.7& 38.4 \\
AdaTAD \cite{liu2024end}& $\usym{2713}$ &11.0G &VideoMAE-L& 87.7& 84.1& 76.7& 66.4& 52.4& 73.5& VideoMAE-L & 57.7& 40.6& 10.1& 39.2\\
AdaTAD \cite{liu2024end} & $\usym{2713}$ & 19.2G& VideoMAE-H & 88.9 &85.3& 78.6 &66.9& 52.5& 74.4& VideoMAE-H & 58.0& 40.6& 9.8& 39.4 \\
\rowcolor{gray!25} \textbf{Our MambaTAD} & $\usym{2713}$ &\textbf{9.3G}& VideoMAE-L & 88.1& 84.1& 78.0& \textbf{67.8}& \textbf{53.3} & 74.3 &VideoMAE-H &63.3&44.4&11.2&43.1 \\
\rowcolor{gray!25} \textbf{Our MambaTAD} & $\usym{2713}$ &\textbf{16.7G}& VideoMAE-H & \textbf{88.6}& \textbf{85.2}& \textbf{78.8}& 67.7& \textbf{53.3} & \textbf{74.7} &VideoMAEv2-G &\textbf{64.2} &\textbf{45.0}&\textbf{11.3}&\textbf{43.8} \\
\bottomrule
\end{tabular}
\label{tab:e2e}
\end{table*}

\begin{table}[t]  
\centering
\caption{Comparison with different multi-label TAD methods on the MultiTHUMOS. 
The average mAP is computed over [0.1:0.9:0.1].
}
\setlength{\tabcolsep}{5.0pt}
\begin{tabular}{l|c|c|c|c|c|c}
\toprule
    Method  &E2E& Features & 0.3& 0.5 & 0.7 & Avg. \\
\midrule
BSN \cite{lin2018bsn}+P-GCN\cite{zeng2019graph}&$\usym{2717}$&I3D  & 16.7& 8.5& - & 10.0 \\
AFSD &$\usym{2717}$&I3D  &23.6 &14.0 & - & 14.7 \\
MS-TCT \cite{dai2022ms}&$\usym{2717}$&I3D&  -&-&-& 16.2 \\ 
PointTAD \cite{tan2022pointtad}&$\usym{2717}$&I3D  & 35.8& 24.9& 12.0& 23.5 \\ 
TadTR &$\usym{2717}$&I3D  &41.1 &29.1& - & 27.4 \\
ActionFormer&$\usym{2717}$&I3D   &44.5 &32.4 & 15.0&  28.6 \\
TemporalMaxer &$\usym{2717}$&I3D & - & 33.4 & 17.4  &29.9 \\
TriDet &$\usym{2717}$&I3D  & - & 34.3& -& 30.7 \\
DualDERT&$\usym{2717}$&I3D   &47.4& 35.2&20.2& 32.6 \\
ADSFormer &$\usym{2717}$& I3D  & \textbf{52.2} & \textbf{40.6} & 21.9 & 35.6 \\
\rowcolor{gray!25} \textbf{Our MambaTAD} &$\usym{2717}$&I3D  & 50.2 & 40.4& \textbf{24.1} & \textbf{35.9} \\
\midrule
ActionFormer*&$\usym{2717}$&VM2&60.8&48.7&28.0&42.3\\
TriDet &$\usym{2717}$& VM2 & -&42.7&24.3&37.5\\
ADSFormer&$\usym{2717}$& VM2 &61.3&51.0& \textbf{31.7} &44.1 \\
ActionFormer*&$\usym{2717}$&IV6&62.6&50.9&29.9&44.0\\
\rowcolor{gray!25} \textbf{Our MambaTAD} &$\usym{2717}$&IV6& \textbf{63.9} &\textbf{52.3}&31.4&\textbf{45.4}\\
\midrule
\rowcolor{gray!25} \textbf{Our MambaTAD} &$\usym{2713}$&VM-S& 59.2&47.7&29.7&42.2\\
\rowcolor{gray!25} \textbf{Our MambaTAD} &$\usym{2713}$&VM-B& 60.1&49.2&31.0&43.1\\
\rowcolor{gray!25} \textbf{Our MambaTAD} &$\usym{2713}$&VM-L& 62.9&52.1&32.8&45.3\\
\rowcolor{gray!25} \textbf{Our MambaTAD} &$\usym{2713}$&VM-H& \textbf{64.2} &\textbf{53.3}&\textbf{34.7}&\textbf{46.6}\\
\bottomrule
\end{tabular}
\label{MultiThumos}
\end{table}

\subsection{Comparison With the State of the Art Methods}
\noindent\textbf{THUMOS14 \& ActivityNet-1.3 with off-the-shelf features.}
Firstly, we compare MambaTAD with leading TAD methods on THUMOS14 and ActivityNet-1.3, as shown in Table \ref{tab:stand_comparison}. Typically, TAD models achieve better performance with advanced backbones and optical flow, as evidenced by the second part of the table. Our evaluation encompasses a wide range of models, including CNNs, GCNs, Transformer-based, Diffusion-based, and Mamba-based architectures.
For a fair comparison, we provide the results of MambaTAD using traditional features in the first part, highlighting its strong adaptability and compatibility. Specifically, MambaTAD surpasses the previous SOTA by 0.5\% on THUMOS14 with I3D \cite{carreira2017quo} features and 1.2\% on ActivityNet-1.3 using R(2+1)D \cite{alwassel2021tsp} features. In the second part, MambaTAD achieves an average mAP of 73.9\% on THUMOS14, outperforming the previous SOTA by 1.2\%. For ActivityNet-1.3, despite its large scale and high diversity leading to lower absolute scores, our method attains an average mAP of 42.8\%, exceeding the previous SOTA by 0.8\%.

\noindent\textbf{THUMOS14 \& ActivityNet-1.3 under end-to-end setting.}
As shown in Table \ref{tab:e2e}, our proposed MambaTAD achieves SOTA performance across both datasets with lower memory usage, demonstrating its superior capability in temporal action detection. 
On THUMOS14, MambaTAD with VideoMAE-Huge as the backbone achieves an average mAP of 74.7\%, outperforming the previous best method (AdaTAD with VideoMAE-Huge) by 0.4\% while using less memory (2.5G less). Notably, even with the lighter VideoMAE-Large backbone, MambaTAD attains 74.3\%, surpassing most existing approaches while maintaining a significantly lower memory footprint (9.3G vs. 19.2G for AdaTAD with VideoMAE-Huge). 
On ActivityNet-1.3, MambaTAD reaches an average mAP of 43.1\% and 43.8\% with the VideoMAE-Huge and VideoMAEv2-Giant~\cite{wang2023videomae} backbone, respectively. It not only significantly surpasses AdaTAD under the same backbone (+3.7\%), but also outperforms the non-end-to-end performance of our MambaTAD with more expressive features InternVideo6B (+1.0\%), demonstrating its efficiency and adaptability across different models and scales. 

\noindent\textbf{Results on MultiThumos.}
MambaTAD also achieves SOTA performance on MultiThumos, excelling in complex, densely labeled action scenarios. With I3D features, it surpasses prior non-end-to-end methods, achieving 35.9\% mAP, outperforming the previous SOTA (ADSFormer, 35.6\%), and demonstrating robust localization.  
Using InternVideo-6B, MambaTAD sets a new record with 45.4\%, increased by 1.4\% compared with the strongest baseline (ActionFormer with InternVideo-6B, 44.0\%). 
Under the end-to-end setting, MambaTAD further improves to 46.6\% mAP with VideoMAE-Huge, the highest reported performance on MultiThumos. It consistently outperforms prior models across all IoUs and remains competitive even with lighter VideoMAE variants, highlighting its adaptability. 

\begin{table}[t]
\centering
\caption{Comparison on the FineAction dataset. 
The average mAP is calculated over [0.5:0.95:0.05].
}
\setlength{\tabcolsep}{5.0pt}
\begin{tabular}{l|c|c|c|c|c}
\toprule
    Method & Features & 0.5 & 0.75 & 0.95 & Avg. \\
\midrule
InternVideo & IV & - & - & - & 17.6 \\
ActionFormer & VM2-g &29.1& 17.7& 5.1& 18.2 \\
ViT-TAD & ViT-B & 32.6 & 15.9 & 2.7 & 17.2 \\
LFAF & X-CLIP & 36.9 & 21.3& 4.5 & 22.2 \\
DyFADet & VM2-g & 37.1 & 23.7 & 5.9 & 23.8 \\
ActionFormer* & IV6B & 43.1 & 27.1 & 5.32 & 27.2 \\
InternVideo2 & IV1B &-&-&-& 27.2 \\
InternVideo2 & IV6B &-&-&-& 27.7 \\
VideoMambaSuite & IV1B &45.4 &28.8 &6.8 & 29.0 \\
\rowcolor{gray!25} \textbf{Our MambaTAD} & IV1B & \textbf{45.6} & \textbf{29.6} & \textbf{5.8} & \textbf{29.4} \\
\bottomrule
\end{tabular}
\label{FineAction}
\end{table}

\begin{table}[t]  
\centering
\caption{Comparison on the HACS dataset. 
The average mAP is calculated over [0.5:0.95:0.05].
}
\setlength{\tabcolsep}{5.0pt}
\begin{tabular}{l|c|c|c|c|c}
\toprule
    Method & Features & 0.5 & 0.75 & 0.95 & Avg. \\
\midrule
TadTR & I3D & 47.1& 32.1& 10.9& 32.1 \\
ActionFormer & SlowFast & 54.9 & 36.9 & 9.5 & 36.4 \\
TriDet& I3D & 54.5 & 36.8 & 11.5 & 36.8 \\
TriDet & SlowFast & 56.7 & 39.3 & 11.7 & 38.6 \\
TCANet & SlowFast & 56.7 & 41.1 & 12.2 & 39.8 \\
InternVideo2 & IV6B & - & -&- &43.3 \\
DyFADet& VM2-g & 64.0 & 44.8 & \textbf{14.1} & 44.3 \\
VideoMambaSuite & IV6B &64.0 &45.7 &13.3 & 44.5 \\
\rowcolor{gray!25} \textbf{Our MambaTAD} & IV6B & \textbf{64.1} & \textbf{46.0} & \textbf{14.1} & \textbf{44.9} \\
\bottomrule
\end{tabular}
\label{HACS}
\end{table}

\noindent\textbf{Results on FineAction and HACS. }
We conducted experiments using various methods and feature sets on FineAction, as detailed in Table \ref{FineAction}. Each method's features are specified, providing a performance comparison. Our method achieves SOTA performance with an average mAP of 29.4\%, outperforming other methods. Despite using IV1B, our method excels over larger parameter features like IV6B, highlighting its effectiveness. 

Lastly, we report the performance of MambaTAD and the previous SOTA methods on HACS in Table \ref{HACS}. Our method achieves an average mAP of 44.9\%, outperforming VideoMambaSuite (+0.4\%). 
The best performance consistently on five datasets with different settings demonstrates the effectiveness and generalization of our approach.

\textbf{Discussion. }
    Across all evaluated datasets, our method consistently outperforms recent state-of-the-art approaches under both non-end-to-end and end-to-end settings. These consistent improvements observed when using the same backbones as competing methods indicate that the proposed architecture provides strong generalization ability and substantial intrinsic contribution to temporal action detection, independent of the backbone employed.

\begin{table*}[t]
    \centering
    \caption{Computational complexity on TAD methods. The best is highlighted in \textbf{bold}, while the second-ranked is \underline{underlined}.}
    \setlength{\tabcolsep}{10.0pt}
    \begin{tabular}{l|c|c|c|c|c|c|c|c|c}
         \toprule
\multirow{2}{*}{Method} & \multirow{2}{*}{E2E}& \multicolumn{4}{c|}{THUMOS14} & \multicolumn{4}{c}{ActivityNet-1.3} \\
\cmidrule{3-10}
 && Backbone&\# params & FLOPs & Avg. & Backbone&\# params & FLOPs & Avg. \\
\midrule
ActionFormer & $\usym{2717}$&I3D& 29.2M & 45.2G & 66.8 &R(2+1)D&6.9M &1.7G & 36.6 \\
TriDet & $\usym{2717}$&I3D& 17.3M & 43.9G & 69.3 &R(2+1)D& 14.0M& 50.6G& 36.6 \\
DyFADet & $\usym{2717}$&VM2-g& 27.6M&45.8G& 71.1 &R(2+1)D& 20.3M& 107.2G& 38.5 \\
InternVideo2 & $\usym{2717}$&IV6B&34.2M&63.6G&\underline{72.0} &IV6B& 9.0M&2.5G&\underline{41.2} \\
\rowcolor{gray!25}\textbf{Our MambaTAD} & $\usym{2717}$ & I3D & \textbf{10.4M} & \textbf{17.8G} & 69.9& R(2+1)D & \textbf{1.4M} & \textbf{0.8G} & 40.2\\
\rowcolor{gray!25}\textbf{Our MambaTAD} & $\usym{2717}$& IV6B&\underline{12.2M} & \underline{19.7G} & \textbf{73.9} &IV6B & \underline{3.5M}& \underline{1.6G} & \textbf{42.8} \\
\midrule
ViT-TAD & $\usym{2713}$ &ViT-B& 131.3M & 866.8G & 69.5 & ViT-B&-&-&37.4\\
AdaTAD & $\usym{2713}$ & VideoMAE-B & \underline{38.4M} & \underline{441.3G} & 71.5 & VideoMAE-H &\underline{89.8M} &\underline{3160.3G}& 39.4 \\
AdaTAD & $\usym{2713}$ & VideoMAE-L &67.2M & 1532.1G & \underline{73.5} & VideoMAEv2-G & 132.0M&5781.0G& 39.8 \\

\rowcolor{gray!25}\textbf{Our MambaTAD} & $\usym{2713}$ & VideoMAE-B & \textbf{19.1M} & \textbf{422.2G} & 70.3 &VideoMAE-H &\textbf{80.9M}&\textbf{3034.6G}&\underline{43.1} \\
\rowcolor{gray!25}\textbf{Our MambaTAD} & $\usym{2713}$ &VideoMAE-L & 46.7M&1469.6G & \textbf{74.3} & VideoMAEv2-G &121.4M&5550.9G&\textbf{43.8}\\
\bottomrule
    \end{tabular}
    \label{tab:com}
\end{table*}
\subsection{Qualitative Evaluation}
\begin{figure*}[t]
    \centering
    \subfloat[]{\includegraphics[width=0.50\textwidth]{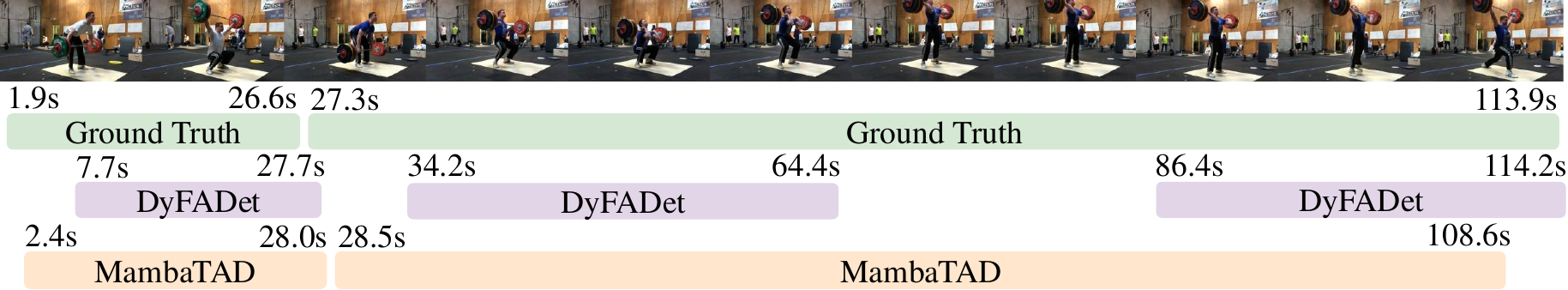} 
        \label{fig:gs}}
    \hfil
    \subfloat[]{\includegraphics[width=0.48\textwidth]{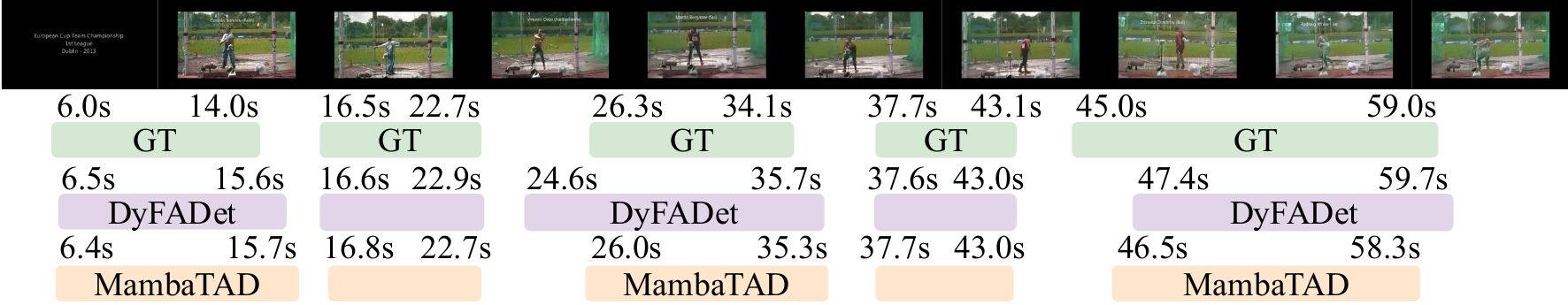} 
        \label{fig:ht}}
    \vspace{-3mm}
    \hfil
    \subfloat[]{\includegraphics[width=0.48\textwidth]{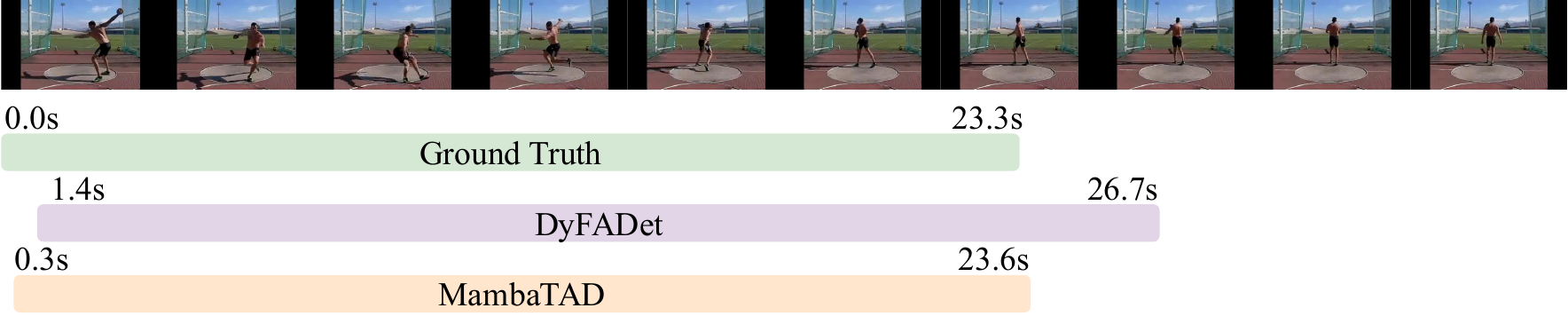} 
        \label{fig:td}}
        \hfil
    \subfloat[]{
        \includegraphics[width=0.49\textwidth]{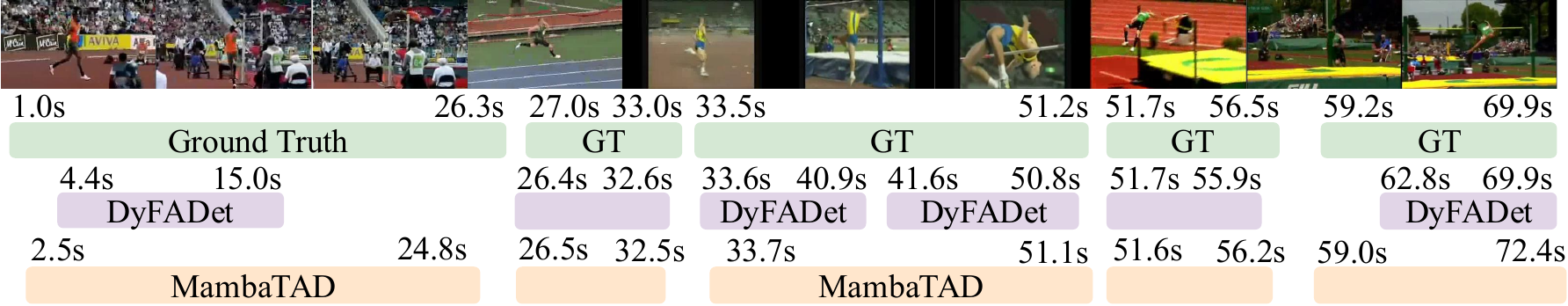}
        \label{fig:hj}}
    \caption{Visualized detection results on the THUMOS14 for action (a) Clean and Jerk; (b) Hammer Throw (occlusion case, best view via zoom-in); (c) Throw Discus; (d) High Jump. }
    \label{fig:vis}
\end{figure*}
Figure. \ref{fig:vis} presents the visual example of the detection results on the THUMOS14 test set with the highest confidence score. It is evident that prior methodologies are unable to fully identify the action ``Clean and Jerk''. Most competing approaches fail to recognize the second instance due to insufficient global context awareness, thus not discerning that the second prolonged movement is a slow-motion replay of the entire lift. In contrast, our method demonstrates superior accuracy due to its robust long-term modeling capability provided by DMBSS and the strategic design of the global detection head.
In Fig. \ref{fig:ht}, almost every frame is heavily covered by a green net, yet our model is still able to accurately detect the action without being affected. This qualitative evidence demonstrates the robustness of MambaTAD to occlusion. Fig.\ref{fig:td} illustrates a simpler class of videos, where only a single action instance is present in the video. It is evident that the previous method exhibits significant error in recognizing the end position, whereas our method more accurately identifies this boundary. This demonstrates that our approach does not inherently favor the recognition of longer instances but instead captures fine-grained boundary features while maintaining global awareness, leading to more precise boundary predictions.
Fig.\ref{fig:hj} represents a more complex scenario, specifically one containing multiple action instances of varying lengths. The previous method struggles with recognizing longer instances, particularly in the case of the high jump action, where the movement significantly slows after the jump. This slowdown likely contributes to the failure of the previous method. In contrast, our method effectively manages this challenge by mitigating temporal context decay and maintaining global awareness, enabling it to identify long instances accurately.

\begin{figure*}[t]
    \centering
    \subfloat[]{
    \includegraphics[width=0.48\linewidth]{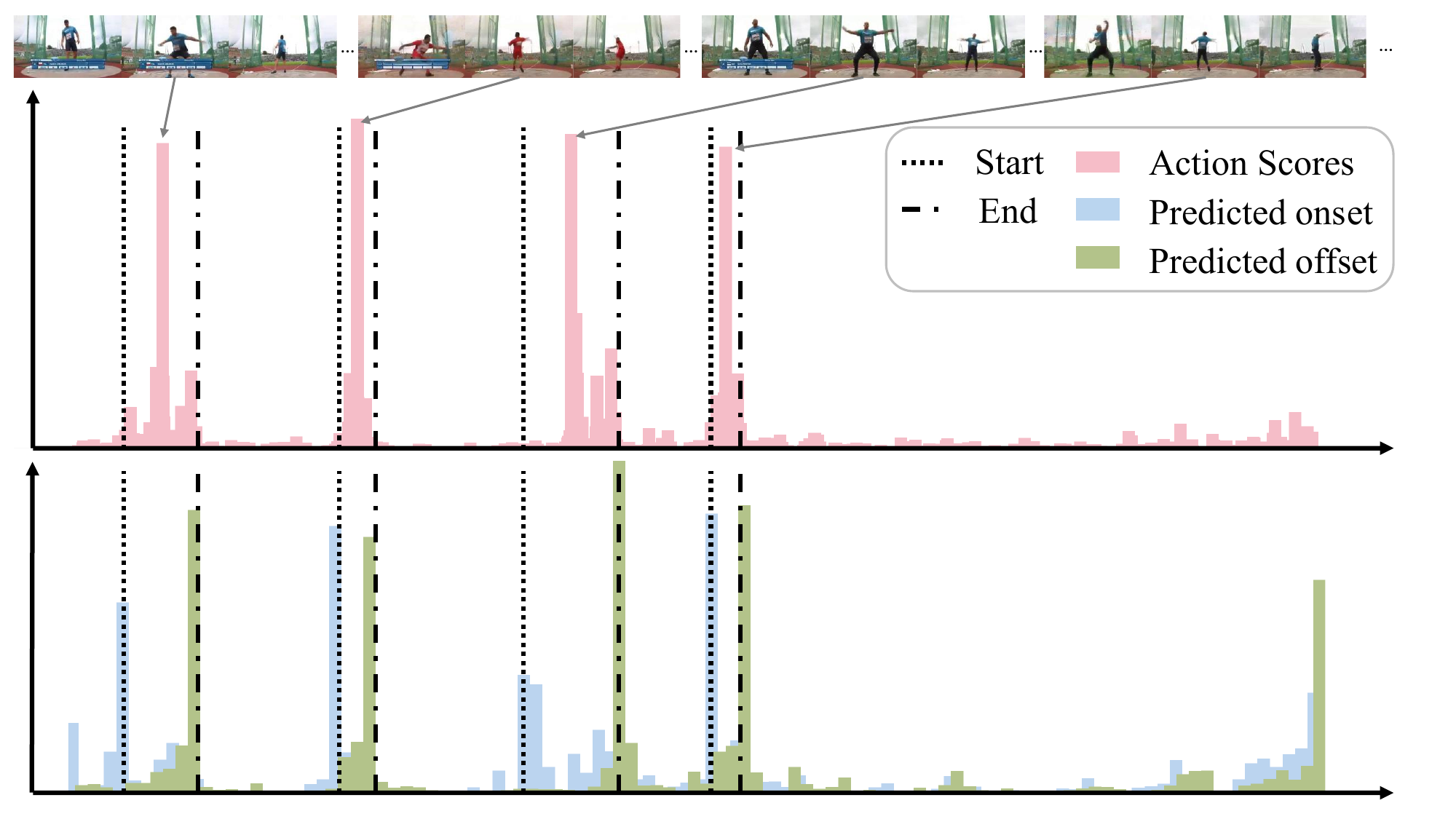}
    \label{fig:stat-td}}
    \hfil
    \subfloat[]{\includegraphics[width=0.48\linewidth]{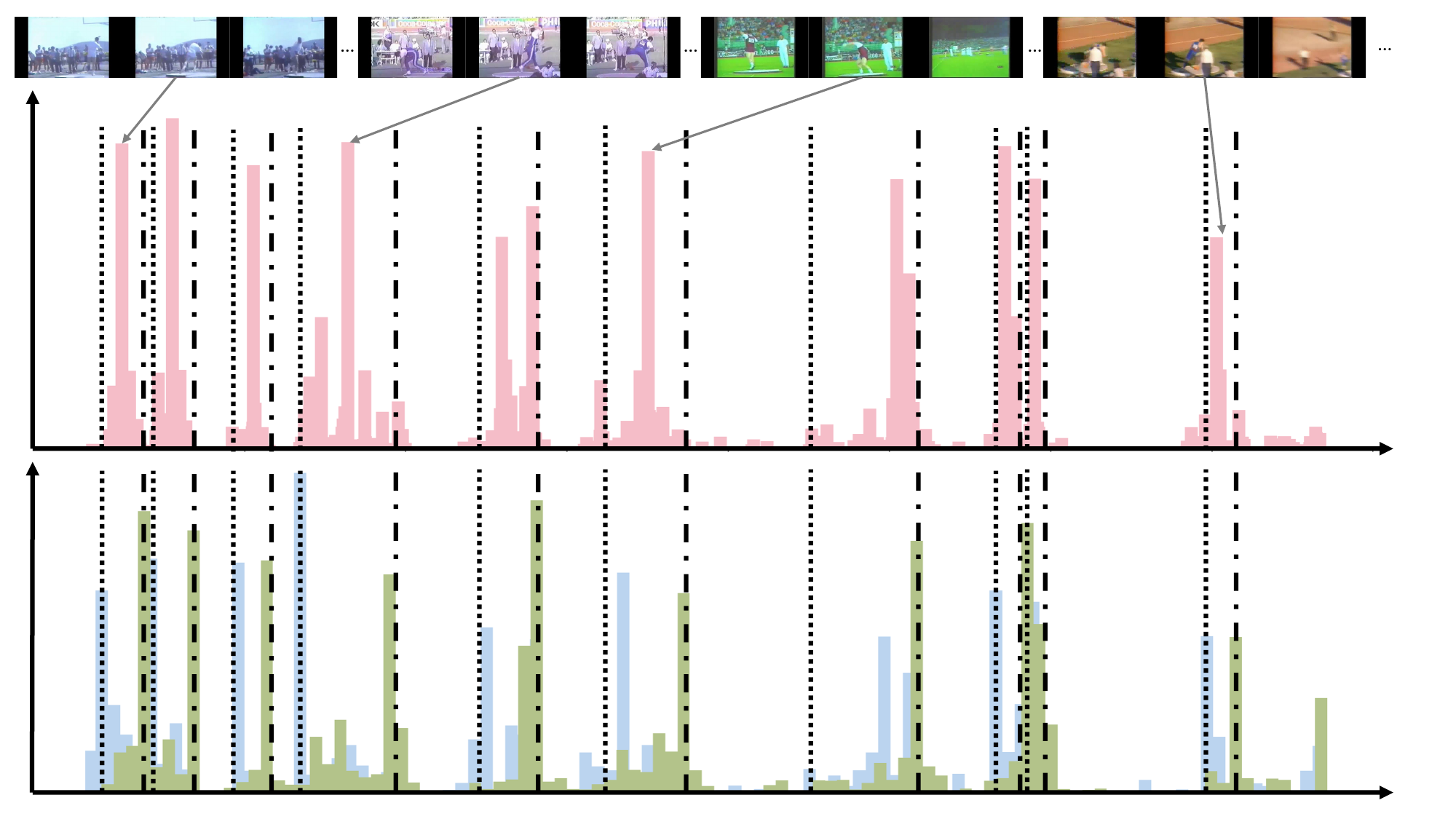} 
        \label{fig:stat-sp}}
    \caption{Additional visualizations of our results are provided. From top to bottom, each item includes: (1) the input video frames, (2) action scores over time, and (3) a histogram of action onsets and offsets, derived by weighting the regression outputs with the corresponding action scores. The square dot line represents the ground truth start position, while the dashed dot line indicates the ground truth end position. This figure is best viewed in color and when zoomed in.}
    \label{vis: stat}
\end{figure*}
\noindent\textbf{Statistical Visualization.}
Further, we present more specific visualizations of our results in Fig. \ref{vis: stat}, including the predicted action scores, and the regression outputs weighted by the action scores (as a weighted histogram). Our model effectively captures video sequences of varying lengths and adapts to action instances of different durations. Even in complex scenarios with multiple action instances, our approach consistently identifies action centers\cite{zhang2022actionformer} and provides accurate boundary predictions. 

\begin{figure}
    \centering
\includegraphics[width=0.95\linewidth]{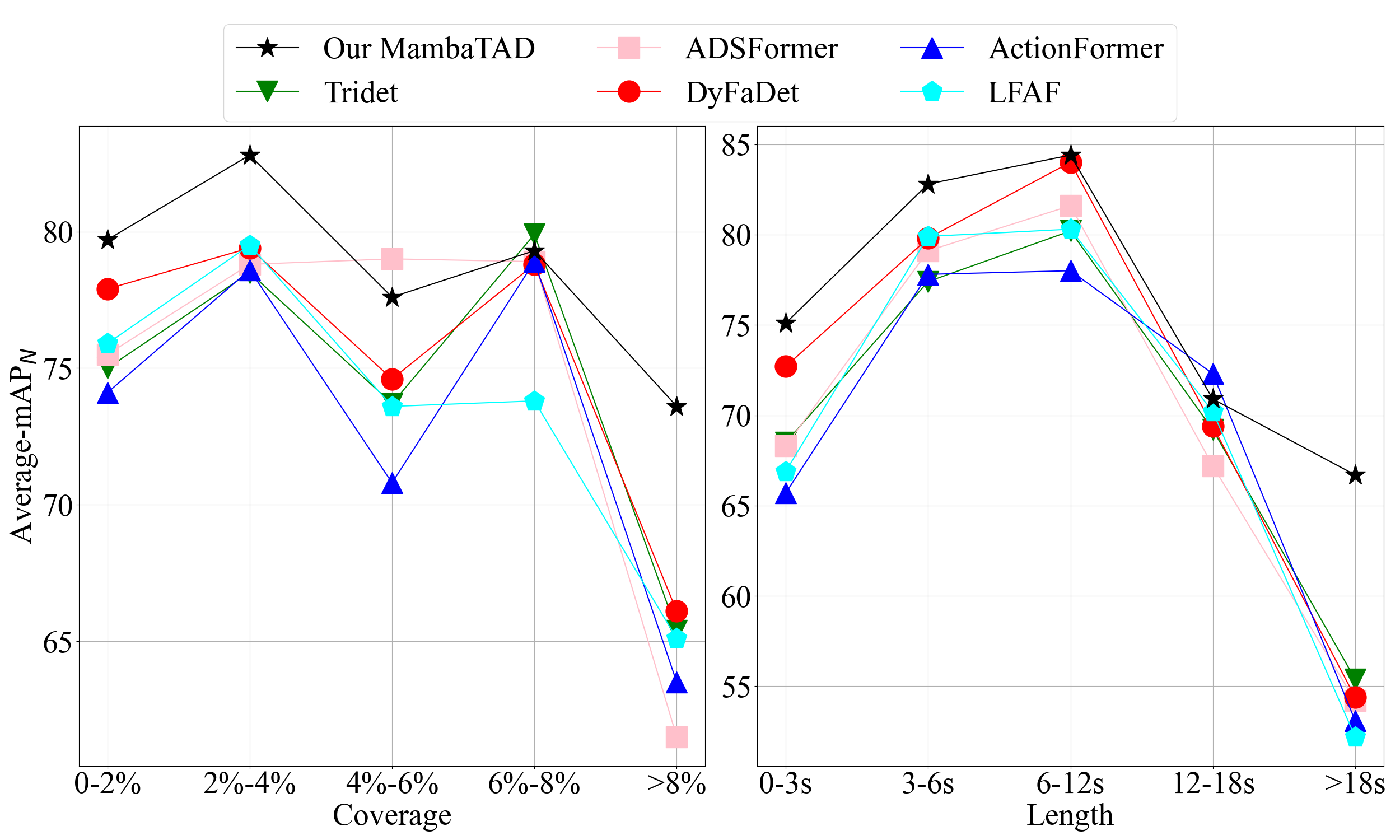}
    \caption{Performance comparison of different TAD methods across varying action instances \textit{coverage} (left) and absolute action \textit{length} (right) on Thumos14. }
    \label{dura}
\end{figure}
\subsection{Robustness Across Varying Action Coverage and Length} \label{sec:sens}
Fig. \ref{dura} compares TAD methods using \textit{Coverage} and \textit{Length} metrics \cite{alwassel2018diagnosing}. 
MambaTAD consistently performs well across coverage levels, excelling in the $0–4\%$ range for short actions and maintaining stability in the \(>\)8\% category, where others significantly decrease. This suggests strong long-range dependency modeling. 
For long actions (\(>\)18$s$), all methods struggle, but MambaTAD shows a more gradual performance drop, indicating better temporal robustness. These results confirm its effectiveness in handling varying action durations, driven by our DMBSS and global feature fusion head, which improve the modeling of complex, extended instances. 
Moreover, we conduct a detailed sensitivity and error analysis in \textit{Supp.} \ref{app:error}.

\begin{figure}
    \centering
    \includegraphics[width=0.95\linewidth]{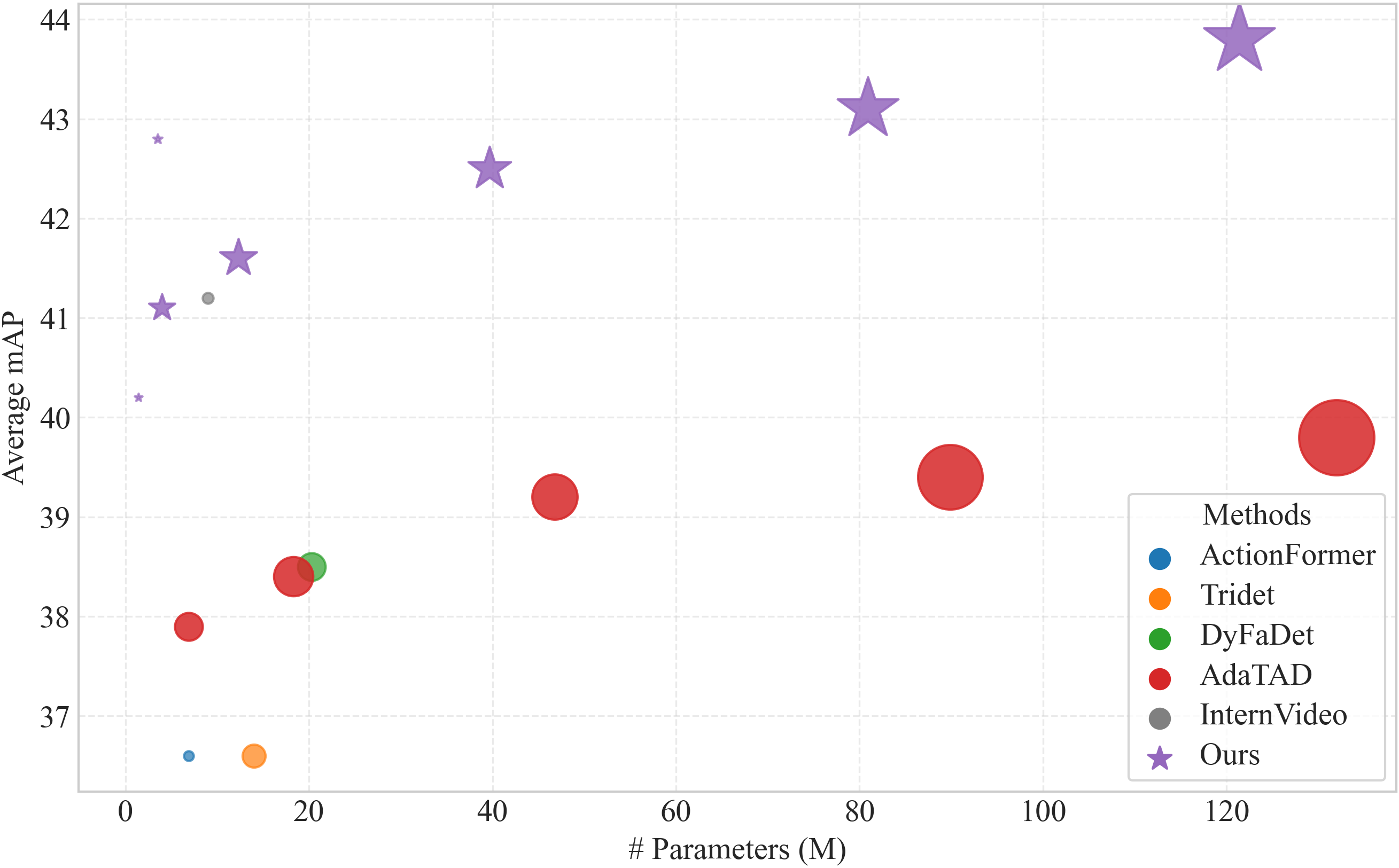}
    \caption{Comparison of model size versus average mAP on ActivityNet-1.3. Bubble size represents FLOPs. Best viewed via zoom-in. }
    \label{fig:com}
\end{figure}
\subsection{Computational Complexity Analysis}
Table \ref{tab:com} presents a comparative computational complexity analysis. Our MambaTAD consistently outperforms other approaches in efficiency and performance, achieving the highest average performance with minimal parameters and lowest FLOPs. 
For instance, 
DyFADet has approximately six times the number of parameters and 67 times the number of FLOPs on ActivityNet-1.3, yet performs worse. This suggests that the efficiency of our method stands in stark contrast to other methods, delivering excellent performance with minimal computational cost. 

Under the end-to-end setting, backbones of different scales primarily determine the model's parameter count and computational cost. However, compared to previous SOTA methods, our approach consistently achieves higher accuracy with fewer parameters and lower FLOPs when using the same backbone. The significant reduction in parameter count is primarily due to the removal of fully connected layers. However, compared to convolution-based methods, the advantage of SSM-based approaches is less pronounced, resulting in similar FLOPs.

In addition, Figure~\ref{fig:com} visualizes the trade-off between model size (number of parameters), computational cost (FLOPs), and accuracy (average mAP) on the ActivityNet-1.3 dataset. Among them, ActionFormer, Tridet, and DyFaDet denote the results obtained using R(2+1)D features under the non-end-to-end setting, while InternVideo refers to the results using IV6B features. AdaTAD represents the end-to-end results with backbones ranging from VideoMAE-Small to VideoMAEv2-Giant. It can be observed that our method achieves superior accuracy and efficiency across all settings. Even with significantly fewer parameters and smaller FLOPs, our model achieves superior mAP values, highlighting its favorable accuracy–efficiency balance and scalability across different model sizes.

\begin{table}[t]
\centering
\caption{Ablation study for the modules in the detector.}
\setlength{\tabcolsep}{10.0pt}
\begin{tabular}{l|c|c|c|c}
\toprule
    Method  & 0.3 & 0.5 & 0.7 & Avg. \\
    \midrule
Baseline  & 86.9 &  75.1  & 50.3 & 72.0 \\
w/ Mamba & 86.3& 76.5& 51.6 & 72.6 \\
w/ DMBSS  & 87.7  & 77.3  & 51.2 & 73.5 \\
w/ Global head  & 87.4  & 77.7  & 52.0 & 73.4 \\
Our MambaTAD & 87.5  & 78.3  & 52.9 & 73.9 \\
\bottomrule
\end{tabular}
\label{table2}
\end{table}

   
\begin{table}[t]
\centering
\caption{Ablation study for different designs of DMBSS. BPS is bidirectional parameter sharing, DM stands for diagonal masks, and DB is the dual branch.}
\setlength{\tabcolsep}{4.5pt}
\begin{tabular}{>{\centering\arraybackslash}p{0.9cm} >{\centering\arraybackslash}p{0.9cm}|>{\centering\arraybackslash}p{0.9cm} >{\centering\arraybackslash}p{0.9cm}|c|c|c|c}
\toprule
BPS & DM& DB & DM  & 0.3 & 0.5 & 0.7 & Avg. \\
    \midrule
 $\usym{2713}$&& & & 86.9 &  76.9  & 52.9 & 72.7 \\
 &&$\usym{2713}$& &86.2& 77.3& 52.3& 73.3 \\
 $\usym{2713}$&&$\usym{2713}$&  & 87.4  & 77.3  & 51.1 & 73.1 \\
$\usym{2713}$&$\usym{2713}$ && &88.0& 77.4& 52.0 & 73.6\\
&&$\usym{2713}$&$\usym{2713}$ & 86.8 & 76.9 & 52.5 & 73.1 \\
$\usym{2713}$&&$\usym{2713}$&$\usym{2713}$ & 87.2 & 77.1 & 52.0 & 73.4 \\
$\usym{2713}$&$\usym{2713}$&$\usym{2713}$ & & 87.4  & 78.0  & 52.0 & 73.6 \\
$\usym{2713}$&$\usym{2713}$&$\usym{2713}$&$\usym{2713}$  & 87.5 &  78.3  & 52.9 & 73.9 \\

\bottomrule
\end{tabular}
   
\label{tab:DMBSS}
\end{table}

\begin{table}[t]
\centering
\caption{Ablation study for adapters of the framework on Thumos14. `AF' stands for ActionFormer. }
\setlength{\tabcolsep}{5pt}
\begin{tabular}{ccc|cc|ccc|c}
\toprule
\multicolumn{3}{c|}{Adapter} & \multicolumn{2}{c|}{Detector} & \multirow{2}{*}{0.3} & \multirow{2}{*}{0.5} & \multirow{2}{*}{0.7} & \multirow{2}{*}{Avg.} \\ \cmidrule{1-5}
Standard & TIA & Our SSTA & AF & Ours &  &  &  &  \\
\midrule
$\usym{2713}$ &  &  & $\usym{2713}$ &  & 86.9 & 75.3 & 51.5 & 72.4 \\
 & $\usym{2713}$ &  & $\usym{2713}$ &  & 87.7 & 76.7 & 52.4 & 73.5 \\
 &  & $\usym{2713}$ & $\usym{2713}$ &  & 87.6 & 77.4 & 52.2 & 73.3 \\
$\usym{2713}$ &  &  &  & $\usym{2713}$ & 87.0 & 76.8 & 52.6 & 73.3 \\
 & $\usym{2713}$ &  &  & $\usym{2713}$ & 87.2 & 77.0 & 52.1 & 73.2 \\
 &  & $\usym{2713}$ &  & $\usym{2713}$ & 88.1 & 78.0 & 53.3 & 74.3 \\
\bottomrule
\end{tabular}
\label{tab:unified}
\end{table}

\subsection{Ablation Study}
\noindent\textbf{Impact of main components in our detector. }
To demonstrate MambaTAD's effectiveness, we initially used ActionFormer with InternVideo-6B features as a baseline. As shown in Table \ref{table2}, the results of the baseline enhanced with vanilla Mamba are presented in the second row. Then, ablation studies on the THUMOS14 dataset compare the baseline with three variants: one with DMBSS, one with the global fusion head, and the complete detector. Both DMBSS and the global head significantly enhance performance over higher thresholds and average metrics, proving their value. DMBSS improves regression with a larger receptive field and bidirectional processing, while the global head ensures effective global feature integration, aiding in handling instances of varying lengths.

\noindent\textbf{Impact of designs of DMBSS. }\label{sec:ablation_dmbss}
This paper tackles two key challenges in applying Mamba to TAD: (1) temporal context decay in unidirectional processing and (2) diagonal conflicts from integrating two triangle matrices. We address these with bidirectional parameter sharing, a dual-branch approach, and diagonal masking.  
As shown in Table \ref{tab:DMBSS}, bidirectional parameter sharing alone reduces our method to DBM \cite{chen2024video}, while a dual-branch structure without sharing improves mAP to 73.3\%, confirming its effectiveness in mitigating temporal decay. However, combining both leads to a slight accuracy drop due to diagonal conflicts. Resolving these conflicts restores accuracy, adding 0.9\% (first line to fourth). Results further show that diagonal conflicts are minimal without parameter sharing, but masking significantly improves modeling performance. Overall, our approach ensures robust boundary representation by balancing bidirectional processing and conflict resolution.

\noindent\textbf{Impact of the adapter. }
We conduct ablation studies on adapters using the VideoMAE-Large backbone, as shown in Table \ref{tab:unified}. First, employing the standard adapter\cite{houlsby2019parameter} with the ActionFormer~\cite{zhang2022actionformer} detector as backbone results in an average mAP of 72.4\%. Replacing the standard adapter with our state-space temporal adapter (SSTA) improves the average mAP by 0.9\%. Similarly, substituting the ActionFormer detector with our detector also leads to a 0.9\% performance gain. It is worth noting that, compared to Temporal-Informative Adapter (TIA), using our adapter alone or combining TIA with our detector results in a slight performance drop. We attribute this to two possible reasons: first, the reduction in parameter count and computational cost may have affected the model's representational capacity; second, there may be an implicit architectural gap when training Mamba-based components alongside non-Mamba structures. In contrast, when both the adapter and detector are Mamba-based, the performance improves significantly. In the unified end-to-end framework, our end-to-end Mamba achieves a 1.9\% improvement over the baseline, demonstrating the strong compatibility of DMBSS with the detector in an end-to-end setting and highlighting the effectiveness of SSTA.

\begin{table}[t]
    \centering
    \caption{Comparison of experimental results under different state-space models.}
    \begin{tabular}{l|c|ccc|c}
    \toprule
    Method & Mamba block & 0.3 & 0.5 & 0.7 & Avg. \\
    \midrule
    baseline &Mamba& 86.3 & 76.5 & 51.6 & 72.6 \\
    Mamba2~\cite{dao2024transformers} & Mamba2 & 83.8 & 72.6 & 42.8 & 67.9 \\
    Hydra~\cite{hwang2024hydra} & Hydra & 84.4 & 72.2 & 43.3 & 68.1 \\
    VideoMambaPro~\cite{lu2024videomambapro} & VideoMambaPro & 86.6& 76.6& 50.4 &72.3\\
    TimeMamba~\cite{chen2024video} & DBM & 86.5 & 75.9 & 49.9 & 72.7 \\
    Vision Mamba~\cite{DBLP:conf/icml/ZhuL0W0W24}& ViM & 86.9 & 76.9 & 52.3 & 73.0 \\
    CausalTAD~\cite{liu2024harnessing} & DBM & 87.4 & 77.6 & 50.8 & 73.1 \\
    \midrule
    Our MambaTAD & DMBSS & \textbf{87.5} & \textbf{78.3} & \textbf{52.9} & \textbf{73.9} \\
    \bottomrule
    \end{tabular}
    \label{tab:2}
\end{table}
\noindent\textbf{Analysis of different SSM-based models. }
As shown in Table~\ref{tab:2}, we conducted experiments with representative state-space models. The results reveal that different Mamba blocks exhibit clear performance gaps under the TAD setting. For example, while DBM-based models (TimeMamba and CausalTAD) achieve stronger baselines compared to Hydra or Mamba2, our DMBSS consistently outperforms all variants across IoU thresholds, demonstrating its advantage in mitigating temporal context decay through bidirectional modeling and self-element conflict via diagonal masks. This analysis further highlights that the decay of temporal context is a common bottleneck across standard SSM variants in TAD, and our design directly addresses this issue.

\vspace{-1mm}
\subsection{Feature Analysis}
\begin{figure}
    \centering
    \includegraphics[width=0.95\linewidth]{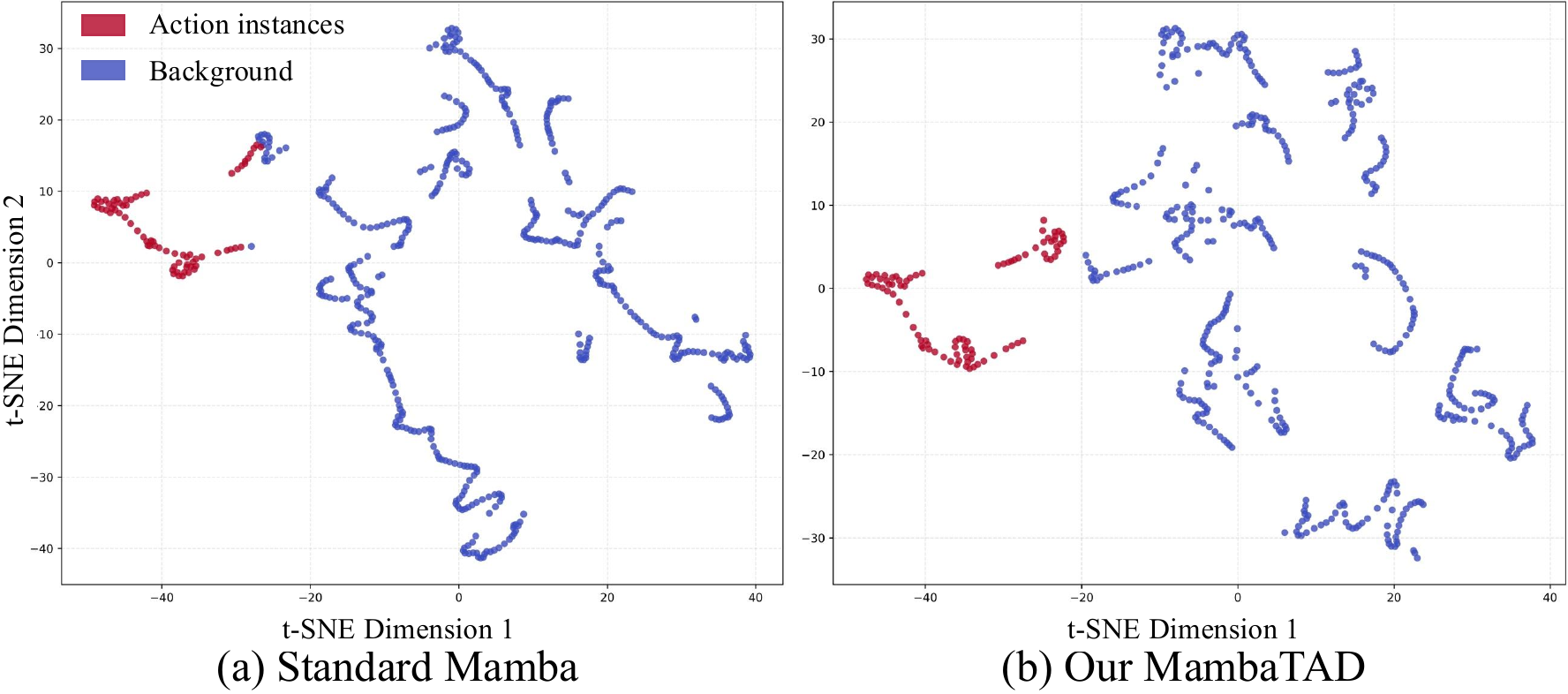}
    \caption{Intermediate feature visualization via t-SNE \cite{van2008visualizing}. (a) the baseline method using the standard Mamba; (b) our proposed MambaTAD}
    \label{fig:feat}
\end{figure}

Figure \ref{fig:feat} presents t-SNE visualizations of intermediate features from the second projection layer (without downsampling), extracted from a randomly selected video containing long-duration actions. Compared to the standard Mamba (Figure \ref{fig:feat}(a)), our proposed MambaTAD (Figure \ref{fig:feat}(b)) yields more compact and well-separated clusters for action instances and background. Notably, MambaTAD demonstrates clearer boundaries between action and non-action segments, indicating its enhanced capability in modeling long-range temporal dependencies and improving action-background discrimination, particularly around temporal boundaries.

\section{Conclusion}
In this paper, we present the MambaTAD, a unified end-to-end one-stage TAD framework leveraging State-Space Models to enhance long-range modeling and global feature detection. We introduce the Diagonal-Masked Bidirectional State-Space (DMBSS) module to improve the model’s capacity for accurately detecting action instances across varying temporal spans. To further refine detection, we propose a global feature fusion head that integrates features from different pyramid levels, ensuring global awareness. Besides, we design a state-space temporal adapter for effective end-to-end TAD. Extensive experiments on multiple benchmarks show that MambaTAD outperforms SOTA models, achieving superior results with fewer parameters and lower computational complexity.

\nocite{wu2025temporal, yu2025towards, Li_Ji_Wu_Li_Qin_Wei_Zimmermann_2024, 10.1145/3581783.3611847,Li_2025_CVPR,li2025secureondevicevideoood, li-etal-2025-treble, yu2024unlearnable, yang2025vidlbeval, li2025personalizedconversationalbenchmarksimulating, liang2023efficient, guo2023elip, ma2025training, zhang2025switch, lu2025pretrain, guo2022unified, guo2025scan, DBLP:journals/corr/abs-2505-05279, yu2025backdoor}
\bibliographystyle{IEEEtran}
\bibliography{mybib}

\vfill
\clearpage
\newpage
\renewcommand{\thesection}{\Alph{section}} 
\setcounter{section}{0}

\section{Implementation details} \label{app:imp}
We use single-stream pre-extracted InternVideo-6B features \cite{wang2024internvideo2} and conduct fair comparisons with I3D \cite{carreira2017quo} (THUMOS14) and R(2+1)D \cite{alwassel2021tsp} (ActivityNet-1.3) features. For end-to-end training, we adopt VideoMAE \cite{tong2022videomae} as the backbone, keeping all but the adapter frozen, with learning rates of $1e-4$ (THUMOS14), $1e-5$ (ActivityNet-1.3), and $4e-4$ (MultiThumos). 
Our detection framework includes two convolutional embedding layers (except MultiThumos) and is optimized with AdamW \cite{loshchilov2017decoupled}. The DMBSS kernel size is 4. Projection layer output dimensions are 512 (THUMOS14, MultiThumos, FineAction) and 256 (ActivityNet-1.3, HACS). We use 7 DMBSS projection layers for THUMOS14, ActivityNet-1.3, and HACS; 9 for FineAction; and 5 for MultiThumos, which also includes 3 convolutional embedding layers for multi-label scenarios. The global feature fusion head partitions features for classification and regression, following \cite{shi2023tridet} for hyperparameters and bins. Our results are obtained by averaging over three independent runs. All experiments are conducted using PyTorch 2.0 and run on a single NVIDIA RTX A5000 GPU, except for the end-to-end setting on the ActivityNet-1.3 dataset, which utilizes 4 A5000 or A40 GPUs.

\section{Error Analysis}\label{app:error}
In this section, we evaluate the detection results on THUMOS14 using the tool from \cite{alwassel2018diagnosing}. The analysis focuses on three main aspects: sensitivity to varying lengths, False Positives (FP), and False Negatives (FN). Specifically, several characteristic metrics were defined in THUMOS14 in \cite{alwassel2018diagnosing}, including coverage, length, and the number of instances. \textit{Coverage} is divided into five bins: Extra Small (XS: $(0, 0.02]$), Small (S: $(0.02, 0.04]$), Medium (M: $(0.04, 0.06]$), Large (L:$(0.06, 0.08]$), and Extra Large (XL: $(0.08, 1.0]$), and \textit{length} is organized into five groups: Extra Small (XS: $(0, 3]$), Small (S: $(3, 6]$), Medium (M: $(6, 12]$), Long (L: $(12, 18]$), and Extra Long (XL: $> 18$). Besides, \textit{the number of instances} represents the total count of occurrences of the same class within a video. This count is further categorized into four groups: Extra Small (XS: $1$), Small (S: $[2, 40]$), Medium (M: $[40, 80]$), and Large (L: $> 80$).
\subsubsection{Sensitivity analysis.} 
\begin{figure}
    \centering
    \includegraphics[width=0.95\linewidth]{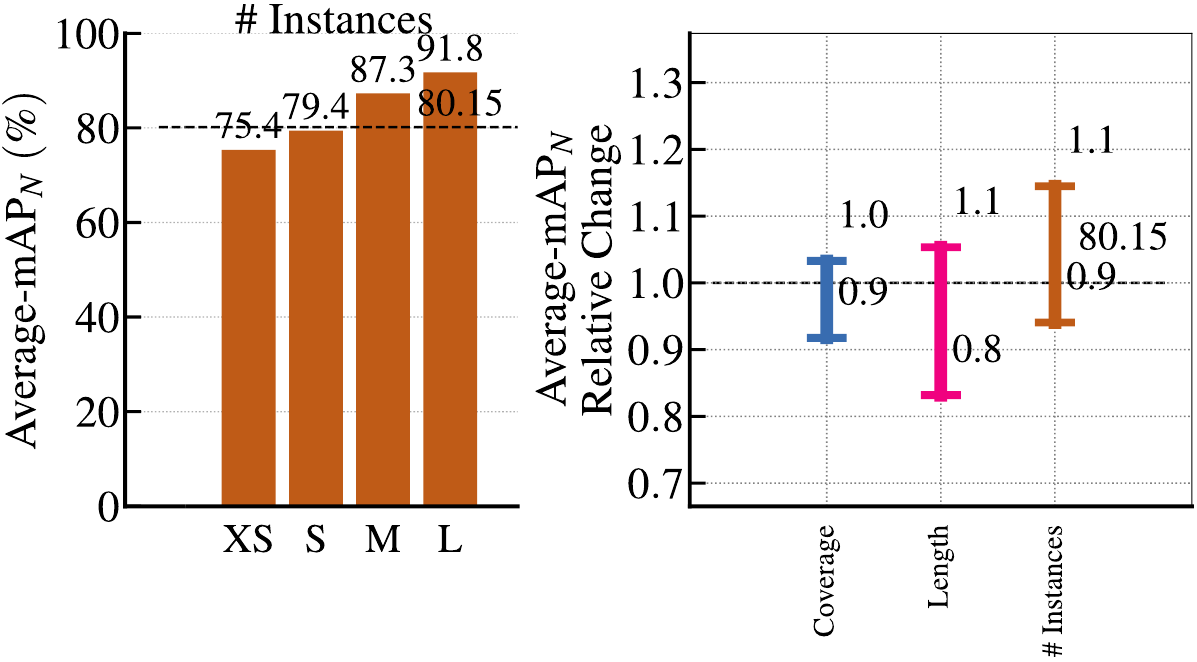}
    \vspace{-4mm}
    \caption{The sensitivity analysis of the detection results on Thumos14.}
    \label{sens}
\end{figure}

In Fig. \ref{sens} (left), the dashed line represents the normalized mAP at tIoU=0.5 across all instances. Sensitivity regarding \textit{coverage} and \textit{length} has been analyzed in Sec \ref{sec:sens}. Further, our method performs slightly below average in the small category but maintains competitive performance in cases where only a single instance of a class appears within a video. This demonstrates strong robustness and generalizability across varying numbers of instances. The right side of Fig. \ref{sens} illustrates the variance of mAP across categories, showing that our method exhibits minimal variation. This indicates that it is relatively insensitive to \textit{coverage}, further reinforcing its stability across different scenarios.

\begin{figure}
    \centering
    \subfloat[]{
        \includegraphics[width=0.95\linewidth]{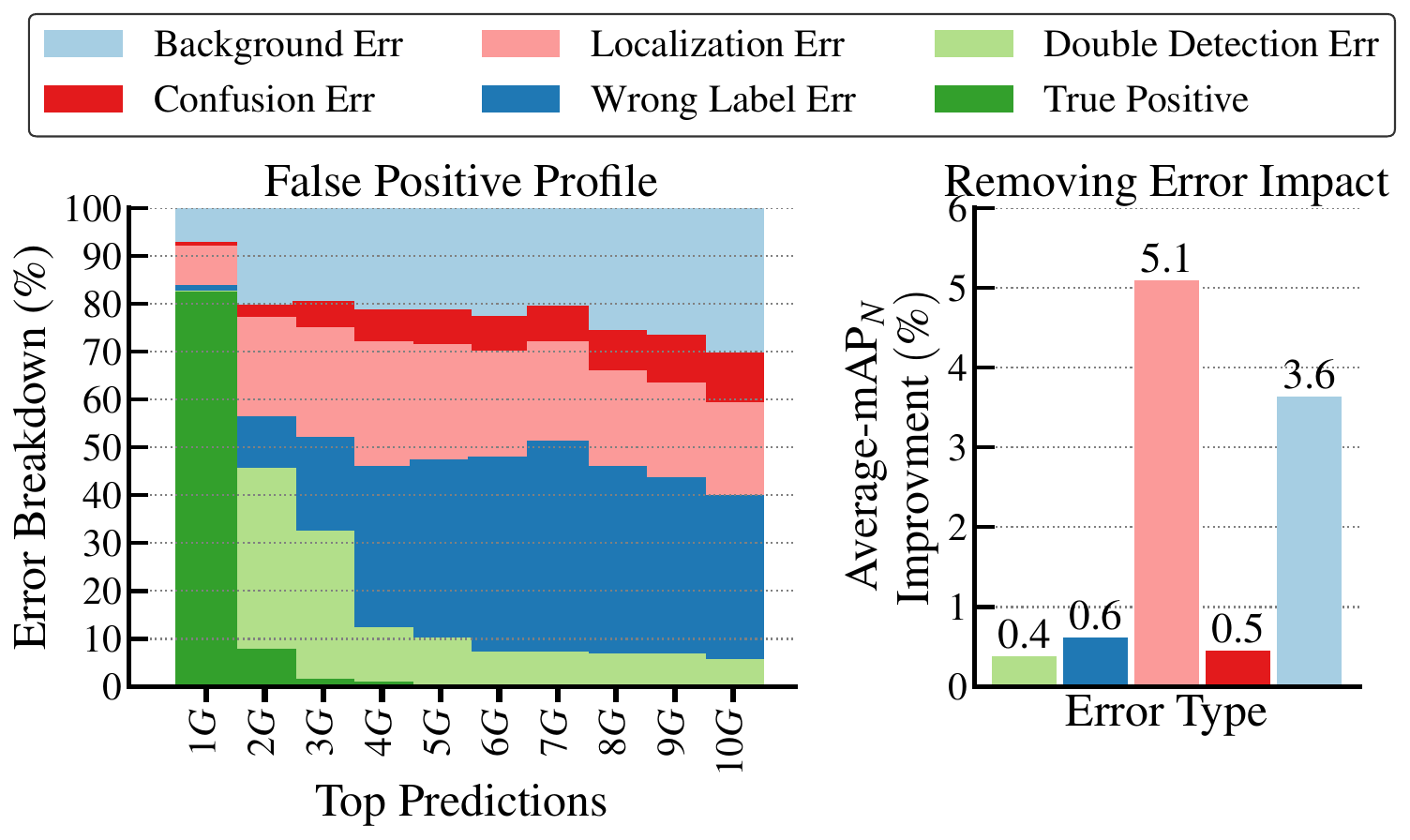} 
        \label{fig:sub1}}
    \hfil
    \subfloat[]{\includegraphics[width=0.95\linewidth]{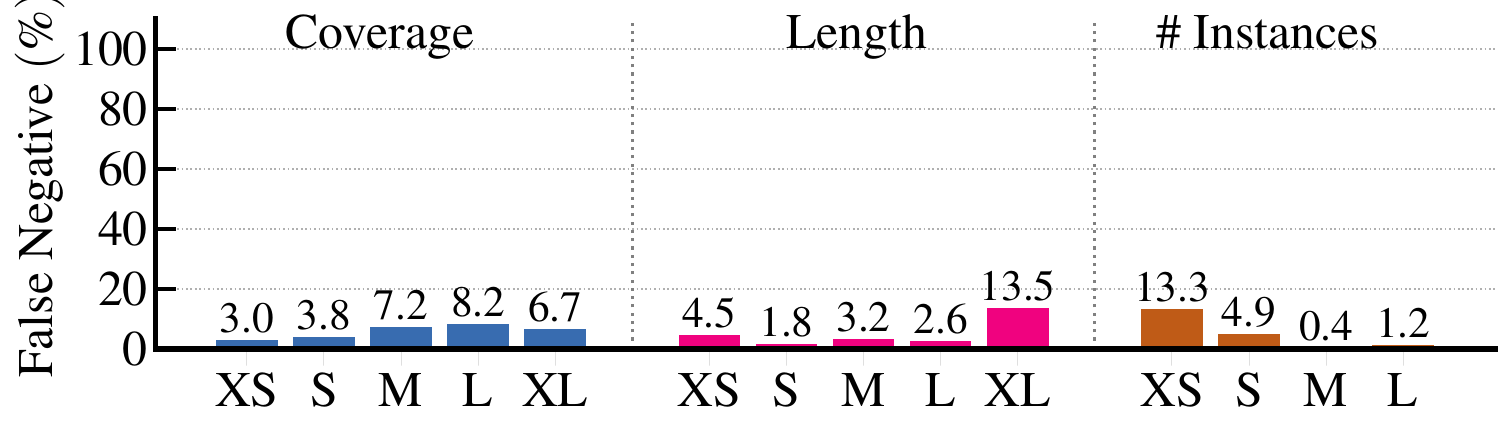} 
        \label{fig:sub2}}
    \caption{The false positive profile and false negative profile. (a) The false positive profile measures the percentage of various common types of detection errors across different Top-K prediction groups. \textit{Right}: The effect of different error types. (b) The false negative profile quantifies the percentage of missed detections across different characteristics metrics. }
    \label{fig:two_images}
\end{figure}
\subsubsection{False positive analysis. }
Fig. \ref{fig:sub1} presents a chart illustrating the distribution of different types of action instances across various k - G values, where G represents the number of ground-truth instances for each action category, and the top k * G predicted instances are displayed for analysis. In the left column for 1G, it is evident that in the top G predictions, true positive instances constitute over 80\% (at IoU=0.5), demonstrating the effectiveness of our method in accurately scoring each instance. Furthermore, on the right, the chart highlights the influence of each error type, with localization errors and background errors (the IoU between prediction and ground truth is significantly below the threshold or equal to zero) still being the primary area of concern. 

\subsubsection{False negative analysis. }
We examine the false negative (miss-detection) rate of our method. As shown in Fig. \ref{fig:sub2}, the longest instances are particularly susceptible to false detection due to their rarity, comprising less than 4\% of the total, which makes it challenging for the model to learn their precise characteristics. Additionally, compared to earlier methods, our approach significantly reduces the false detection rate for instances shorter than medium length, demonstrating its effectiveness in handling fine-grained short instances. Our method shows improved performance on videos with a higher number of instances, indicating better handling of more complex scenarios.

 




\vfill

\end{document}